\documentclass[journal]{IEEEtran}
\usepackage{lineno,hyperref}
\usepackage{times}
\usepackage{epsfig}
\usepackage{graphicx}
\usepackage{subfigure}
\usepackage{amsmath}
\usepackage{amssymb}
\usepackage[usenames,dvipsnames]{xcolor}
\usepackage{tabularx}
\usepackage{multirow}

\ifCLASSINFOpdf
\else
\fi
\begin{document}
%
\title{Enhanced Mixtures of Part Model for Human Pose Estimation}
%
%
%

\author{Wenjuan~Gong,
        Yongzhen~Huang,
        Jordi~Gonz\`alez,
        and~Liang~Wang
\thanks{W. Gong is with the Department
of Computer and Communication Engineering, China University of Petroleum, Qingdao, China, e-mail: wenjuangong@upc.edu.cn.}
\thanks{Y. Huang and W. Liang are with Institute of Automation, Chinese Academy of Science.}
\thanks{J. Gonz\`alez is with Computer Vision Center, Autonomous University of Barcelona.}
}

%
%

\markboth{}
{Gong \MakeLowercase{\textit{et al.}}: An Effective Solution to Double Counting Problem in Human Pose Estimation}
%



\maketitle

\begin{abstract}
   Mixture of parts model has been successfully applied to 2D human pose estimation problem either as explicitly trained body part model or as latent variables for the whole human body model. Mixture of parts model usually utilize tree structure for representing relations between body parts. Tree structures facilitate training and referencing of the model but could not deal with double counting problems, which hinder its applications in 3D pose estimation. While most of work targeted to solve these problems tend to modify the tree models or the optimization target. We incorporate other cues from input features. For example, in surveillance environments, human silhouettes can be extracted relative easily although not flawlessly. In this condition, we can combine extracted human blobs with histogram of gradient feature, which is commonly used in mixture of parts model for training body part templates. The method can be easily extend to other candidate features under our generalized framework. We show 2D body part detection results on a public available dataset: HumanEva dataset. Furthermore, a 2D to 3D pose estimator is trained with Gaussian process regression model and 2D body part detections from the proposed method is fed to the estimator, thus 3D poses are predictable given new 2D body part detections. We also show results of 3D pose estimation on HumanEva dataset. 
\end{abstract}


%
\IEEEpeerreviewmaketitle

\section{Introduction}\label{sec:introduction}

Pose estimation from still images has wide applications in image and video indexing, video surveillance and human computer interaction. For example, online solutions of this problem can be applied for single frame initialization in tracking human poses. Yet pose estimation from still images, that is, 2D body part localization is a difficult problem, due to the fact that human body is highly flexible resulting human poses with high degrees of freedom even in 2D images. 

A state-of-art and currently widely used solution for 2D body part detection is the mixture-of-parts (MoP) method~\cite{cvprYangR11APEMOP}, in which a human body is modeled as a tree structure and body parts are encoded as nodes in the tree. Maximum responses from detection are passed from the leaf nodes to the root. One problem with this solution is the double-counting problem, that is, one detected body part is counted twice for both sides of the human body. In this paper, we tackle the double-counting problem in MoP model with multiple feature inputs. Additional input features are incorporated so we are able to verify body part localization from more feature responses. Compared with the greedy solution in the original MoP method, we are able to solve double-counting problem with a global optimization.

\begin{figure*}
\centering
\includegraphics[scale=0.45]{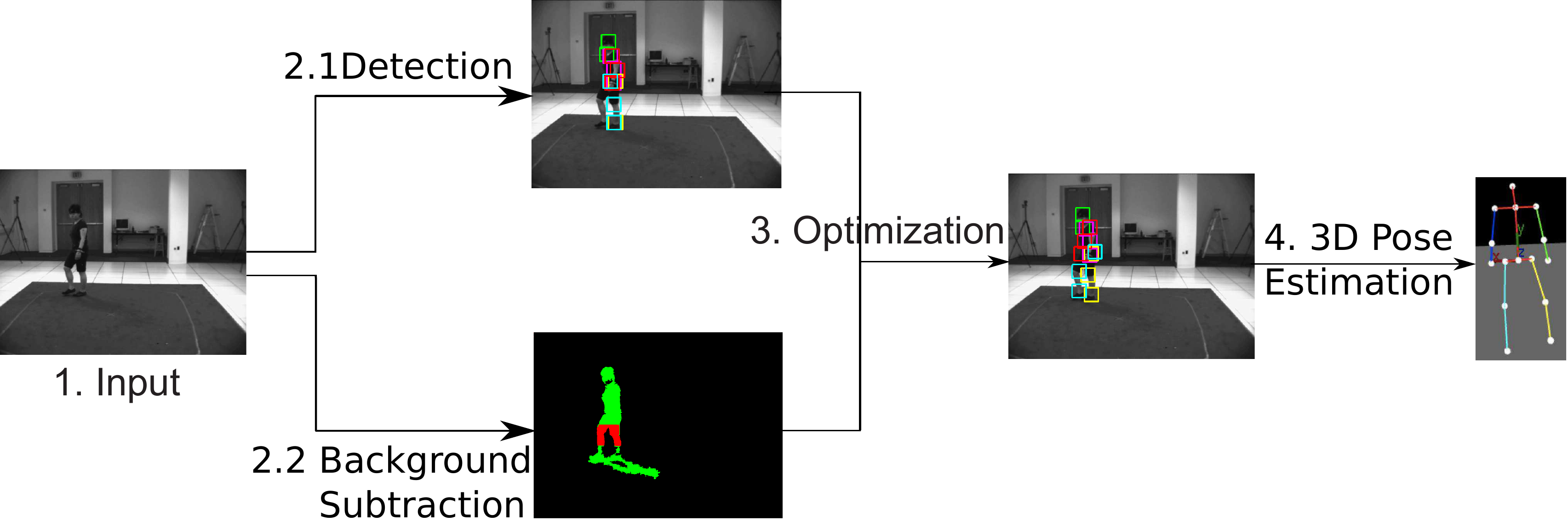}
\caption{2D body part detections and 3D pose estimation based on enhanced Mixtures of Parts method. The pipeline includes several major steps: feature extractions (MoP and background subtraction, in our case), global optimization, and 2D to 3D pose estimation.}
\label{fig:pipelineoverall}
\end{figure*}
With detected 2D body part locations from the proposed method, we are able to predict 3D poses by feeding 2D body part detections to a 2D to 3D pose estimator. For 2D to 3D pose estimator, we choose Gaussian process regression, which has been proved to be effective in modeling non-linear regression problems. We further validate the whole pipeline on a public available dataset for pose estimation: HumanEva dataset. We visualized two types of results: enhanced 2D body part detections and 3D poses estimated from enhanced 2D body part detections. Figure~\ref{fig:pipelineoverall} shows main steps as a pipeline for the whole algorithm. The pipeline includes several major steps: feature extractions (MoP and background subtraction, in our case), global optimization, and 2D to 3D pose estimation. From the input, the original mixtures of part model is trained and applied to detect body part positions. Meanwhile, human blobs extracted with background subtractions are used as another cue in our method. Then in the third step, this two cues are combined with the proposed algorithm.



The generalized framework in our algorithm are able to incorporate multiple features other than human blobs. We extract multiple cues from input images and combine them under the proposed framework. With augmented inputs from multiple features, we are able improve 2D body part localization and solve double counting problem. As in MoP method, a human body is modeled by a tree structure where kinematic constraints between connecting parts are kept. First, feature models of different features are trained separately and the optimization target is modified to reach a global optimization target by incorporating multiple feature cues. The inference of the optimal pose in a test image is also carried out with multiple cues. The bottom up message passing procedure combines multiple feature cues so as to reduce false positive body part detections and the top down back tracing procedure are optimized globally so as to tackle double counting problem.

The contributions of this papers are as followings: by combing multiple cues, we boost 2D body part localizations under a general framework; enhanced 2D body part detectors are validated on a public available dataset: HumanEva dataset; 3D pose estimation is shown as an examplar application of detected 2D poses and this application is also validated on HumanEva dataset. The rest of the paper is organized as following: in section~\ref{sec:rw}, we introduce related works on 2D pose estimation and related works on solving double counting problems; in section~\ref{sec:themethod}, we introduce details of the proposed method including training models from different features and the global optimization taget; section~\ref{sec:results} shows boosted 2D body part localizations by combining multiple feature cues on a standard public available dataset and the examplar application of 3D pose estimation from localized 2D body parts; in section~\ref{sec:conclusions} we conclude the work and discuss about possible future works.

\section{Related Work}\label{sec:rw}
As mentioned in the previous section, human bodies are highly flexible, thus results in a huge amount of possible guesses in the solution space. But human body joints are not completely under no constraints. Models like pictorial structure model~\cite{Felzenszwalb2005PSOR} and tree models~\cite{Xiao2012PCDCPCPE,Wang2008MTMMTMOSCHPE,cvprYangR11APEMOP} are exploited and successfully applied to represent human body models in 2D. These models keep kinematic constraints between connecting body parts. That is, the body parts that are connected physically are also connected in the tree structure. Using tree structures has the advantage of tractable inference of the optimal pose. However, spatial constraints between body parts without direct connections are not incorporated in the tree structure. Due to this reason, the original tree structure cannot deal with occlusion and has the problem of double counting, where an image evidence is counted twice for different body parts.

One important example of tree models is mixture of parts model~\cite{cvprYangR11APEMOP}. MoP defines body parts as an area surrounding body joints and has the advantage in dealing with the foreshortening problem of body limbs caused by viewpoint changes. While the traditional pictorial structure (PS) model~\cite{Felzenszwalb2005PSOR}, which defines body parts as body limbs, needs to deal with foreshortening problem by explicitly training on body limbs of different lengths. Also in MoP model, the orientation of a limb is naturally represented by the connection of detected body joints. While in a traditional PS model, limb orientations need to be learned and detected explicitly. So we choose MoP model as the human body model. A body part in the MoP model is represented as a mixture of several templates, each of which is trained with one subset of samples of this body part. In this way, the trained body part is able to deal with different limb layouts from different poses.

As mentioned in the first paragraph, although the tree structured human model is efficient in training and referencing, it has the double counting problem due to occlusions and lack of constraints between body parts denoted as a node in the tree structure. To deal with these problems, authors in~\cite{Wang2008MTMMTMOSCHPE} propose multiple tree models. The models contain a tree structure to account for kinematic constraints between connected body parts, tree structures for spatial constraints among body parts without direct connections, and tree structures for occluded body parts. Different tree structures are combined with a boosting procedure. Other research also explore the possibility of imposing constraints in the optimization target. For example, authors in~\cite{Xiao2012PCDCPCPE} modify the optimization target and incorporate spatial constraints to deal with double counting problem. In referencing, those poses who violate the spatial constraints will get a comparatively lower score.


\section{The Method}\label{sec:themethod}

Given training images with only one human in each image, we train 2D body part detectors with image patches cropped within bounding boxes surrounding the body parts. For a test image, we localize 2D body part positions with trained detector and optimize the detection with multiple feature cues. We name the detector enhanced MoP model since it is based on the MoP model proposed in~\cite{cvprYangR11APEMOP}. In the following subsection, we are going to split the method into modules and explain in details.

\subsection{Mixture of Parts}
The idea of mixture of parts detector in~\cite{cvprYangR11APEMOP} is to represent a body part with a mixture of several ($5$ or $6$) templates, each represent a different appearance of the corresponding body part. So the body part which has more variances in appearance, for example, elbows and knees, are apt to contain more templates. After cropping the image surrounding the bounding box with a proper size, all samples of a body parts are clustered into several clusters, whose total number is predefined according to the variance of the body part. Training templates are formulated as optimizing parameters in a support vector machine, which is carried out with EQ optimization. Note that the size of the bounding box is a crucial factor in adapting the method to custom data. Considering the different notation in each dataset, body joints might correspond to different position and if the size of the bounding box is defined too big, it might contain information from other body part and if the size is too small, it might be lack of information for identifying the body part or joint.

After training templates for each body part, given a test image, we compute the response of the image with respect to all trained templates by convolution. Then a distance transform~\cite{Felzenszwalb2004EGI} is performed so that the maximum response of the image to the test template is highlighted. Later on, we start from the leaves of the human tree structure (rooted at the head), and pass maximum responses of all mixtures from the child body part to its parent. Thus, when we come to the root node, all the body part nodes contribute by passing messages. The score of the root is considered the final score of the human detection. This tree structure is very effective in referencing but it has problem dealing with double counting problem. In the following subsections, we are going to explain how we are going to enhance the algorithm based on the original MoP model.

\subsection{Enhanced MoP Via Multiple Cues Fusion}

Instead of imposing spatial constraints or modifying tree structure model, we explore the possibility of combining multiple cues from input images. We argue that multiple feature cues provides richer information so that effectively combining multiple cues reduce false positives and ease double counting problem. In our experiments, we consider histogram of gradient (HOG)~\cite{Dalal2005HOG} and human blobs extracted from background subtraction~\cite{Amato08backgroundSubtraction}.

\subsubsection{Formulation of Enhanced Model}

Let us write $I$ for an image, $p_i = (x\; y)$ for the pixel location of part $i$ and $t_i$ for the mixture component of part $i$. We write $i \in \{1,\dots, K\}$, $p_i \in \{1,\dots,L\}$ and $t_i \in \{1,\dots,T\}$. We call $t_i$ the ``type'' of part $i$. For notational convenience, we define the lack of subscript to indicate a set spanned by that subscript (e.g., $t =\{ t_1, \dots, t_K\}$). The kinematic constraints of human body between connected body parts are modeled as following:

{\footnotesize
\begin{equation}
\displaystyle S(t) = \sum_{i\in V}b_i^{t_i} + \sum_{ij \in E} b_{ij}^{t_i,t_j}.
\end{equation}
\par}
The parameter $b_i^{t_i}$ favors particular type assignments for part $i$, while the pairwise parameter $b_{ij}^{t_i,t_j}$ favors particular co-occurrences of part types. We write $G = (V;E)$ for a K-node relational graph whose edges specify which pairs of parts are constrained to have consistent relations.

We can now write the full score associated with a configuration of part types and positions:

{\footnotesize
\begin{equation}
\displaystyle S(t) =  S(t) +  \sum_{i\in V}\omega_i^{t_i}\cdot \phi(I,p_i)  + \sum_{ij \in E} \omega_{ij}^{t_i,t_j}\cdot \psi(p_i-p_j), \label{eq:PMoP}
\end{equation}
\par}
where $\phi(I,p_i)$ is a HoG vector extracted from pixel location $p_i$ in image $I$.  $\psi(p_i-p_j) = [dx\quad dx^2\quad dy\quad dy^2]^T$ , where $dx = x_i - x_j$ and $dy = y_i - y_j$ , the relative location of part $i$ with respect to $j$.

Until now, this is the original MoP model from~\cite{cvprYangR11APEMOP}. Since multiple features are extracted separately, we can train models from each candidate features separately. For example, when we use extracted human blobs as another feature cue. We get the human blob model from background subtraction as following:

{\footnotesize
\begin{equation}
F(p_i) = \left\{
\begin{matrix}
  1, if |E(p_i) - B(p_i)|> threshold,\\
  0, if |E(p_i) - B(p_i)|< threshold,\\
 \end{matrix} \right.
\label{eq:ForegroundHB}
\end{equation}
\par}
where $B(p_i)$ is the background model and can be updated with new added frames in the following way,

{\footnotesize
\begin{equation}
B_{t+1}(p_i)= \alpha* F_{t}(p_i)+ (1 -\alpha)*  B_{t}(p_i).
\end{equation}
\par}
And $\alpha$ is the learning rate. We denote this model as human blob (HB) model.

After training MoP model and HB model separately, given a test image, we need to find the optimal human pose with respect to certain criterion. This criterion should take into account both of the trained models. Since we suppose each image features are extracted separately. We can get the joint probability of matching two models as:

{\footnotesize
\begin{equation}
P(M,H) = P(M|H)\cdot P(H),
\end{equation}
\par}
where $M$ represents the MoP model and $H$ represent the HB model. This probabilities formulation can be easily extended to other image feature cues, given the definition of the model probabilities and conditional probabilities.

In our method, we consider HOG and detected human blobs. We define $P(H)$ for each pixel $m$ as following:

{\footnotesize
\begin{equation}
P_{m}(H) = \left\{
\begin{matrix}
1, \quad if F_{m}=1,\\
o, \quad if F_{m}=0.
\end{matrix}
\right.
\end{equation}
\par}
Since the HOG feature and the human blob feature are extracted separately and thereafter MoP model and HB model are trained separately, $P(M|H)$ equals $P(M)$ which is defined in equation~(\ref{eq:PMoP}). In implementation, we calculate the probability of a certain pixel $m$ belonging to a certain body part by convoluting with image evidence of this pixel with trained body part template.

\subsubsection{Finding Root Positions}

After training MoP model and HB model separately from HOG and human blob features. We can detect human pose from an unseen image by find the optimal human pose. In~\cite{cvprYangR11APEMOP}, with all trained mixtures of parts models, the test image is convoluted with each trained templates. Then starting from the leaves of the tree structure, responses of all body parts are passed to their parent parts. After one pass, all the body parts contribute their score to the root part of the tree structure.

In our combined model, before passing the score from the child node to its parent node, we check if this pixel also confirms with the evidence from human blob detection. If the current pixel belongs to the detected human blob, we keep the current score, otherwise the score is set to a very small value. This procedure guarantee that the final probability is the joint probability of two candidate feature models. The advantage of this procedure is obvious, we can remove some false positives by verifying that the current pixels confirm with both models trained from different image feature cues. So the detected root position is more accurate. After we find the root position of the human, we go through the whole tree to fix each body parts with global optimization.

\subsubsection{Finding Body Part Positions}

From the detected root position of the tree structure, authors in~\cite{cvprYangR11APEMOP} employ a backtracking algorithm to fix all the body part positions. It starts from the root of the tree structure and fix its child node by picking the maximum response from all the child nodes. This method causes the double counting problem. Since each body part is fixed only considering the response of the test image with the trained templates, when sibling body parts (the same body part, but on different sides of the human body, like a left hand and a right hand) resemble each other, the same image patch might be picked repetitively. In this case, the estimated pose is occluded while in the real case it is not.

To solve this problem, we use global optimization to fix each body part positions. In the original MoP model, where there is only HOG feature, optimizing body part position is very time consuming. For example, if the model uses $26$ body parts, and each body parts use $5$ or $6$ mixtures, the minimum number of possible combination for all body parts is $5^25$. This is a huge amount of possible guesses. In our case, where we consider human blob as another feature cues. The possible number of mixtures for each body parts is great reduced due to the constraint. So we can optimize the tree structure globally. The body part positions are optimized to maximize the score of the proposed model for combining multiple cues $S_{MC}$:

{\footnotesize
\begin{equation}
\sum_{i\in V}\omega_i^{t_i}\cdot \phi(I,p_i)\cdot F(p_i)  + \sum_{ij \in E} \omega_{ij}^{t_i,t_j}\cdot \psi(p_i-p_j)\cdot F(p_i)\cdot F(p_j), \label{eq:S_MC}
\end{equation}
\par}
where $F(p_i)$ is the ratio of overlap between the body part $p_i$ and the foreground model define in equation(\ref{eq:ForegroundHB}).

\begin{figure}[!ht]
\centering
\includegraphics[scale=0.25]{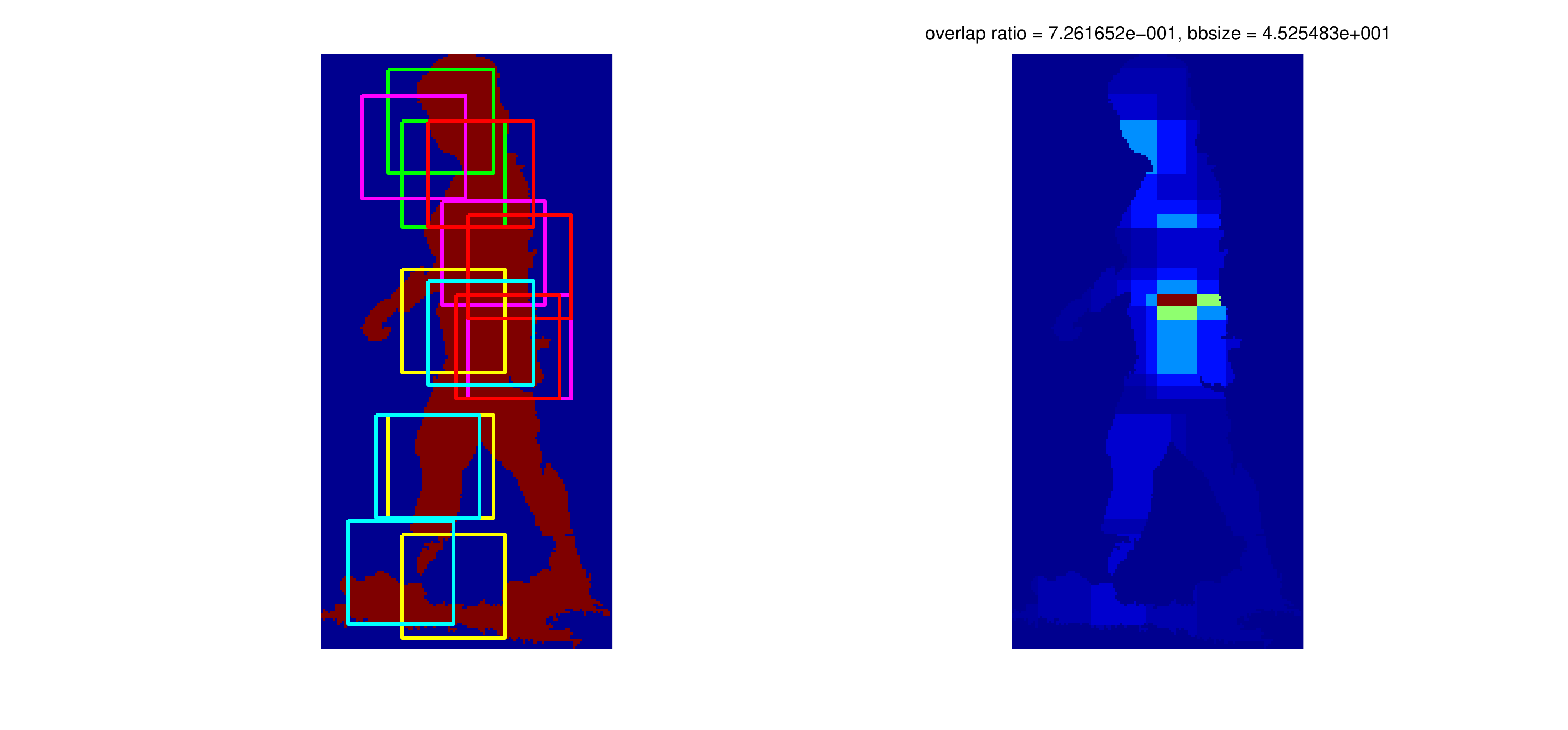}
\caption{Detected body part localization with the Mixture of Parts (MoP) model. The left figure shows the localizations of all body parts with MoP. The right figure shows the overlap between extracted human blobs and all bounding boxes. Color variations in human blobs denotes different number of bounding boxes that are overlapped.}
\end{figure}

\begin{figure}[!ht]
\centering
\includegraphics[scale=0.25]{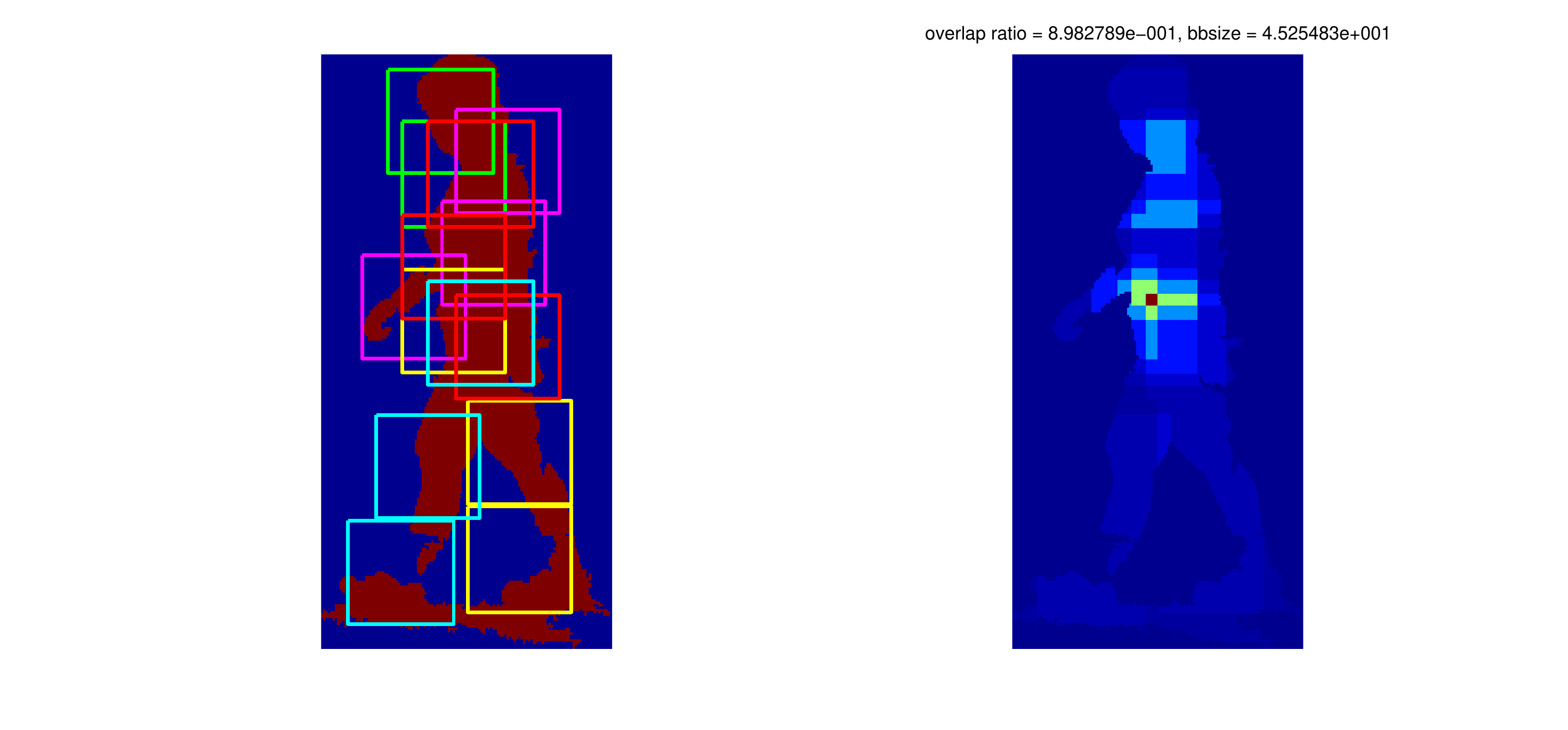}
\caption{Body part localization after optimization (left) and final overlap map between all bounding boxes and the extracted human blob.}
\end{figure}

\subsection{From 2D Parts to 3D Pose Estimation}
The Gaussian process regressor is one of the most widely used regression model for learning 2D to 3D mapping in the pose estimation since it has been proved to be an effective approach for the nonlinear
2D to 3D pose mapping problem~\cite{Gregor09Nonlinear,Sofiane04Nonstationary,Wang08DynamicalModel}. Gaussian Process Regression (GPR) is
considered a model-free framework. GPR defined as a distribution
over functions, extends statistics from data points to
functions. With kernel trick, we can even get rid of the function
definition, and only concentrate on kernel matrix instead. Once
we normalize the training input to have a zero mean, we only
need to define a covariance matrix, that is ,the kernel matrix,
for GPR. Frequently used covariance matrices include squared
exponential covariance function, Mat\'{e}rn covariance function
and so on. In the following subsections, we will explain detailed representations and settings for the Gaussian process regressor used here.

\subsubsection{Definition of Gaussian Process Regression}
According to~\cite{Carl06Gaussian}, Gaussian process is defined as:
\emph{a collection of random} \emph{variables, any finite number of
  which have (consistent) joint} \emph{Gaussian distribution}.  A
Gaussian process is completely specified by its mean function and a
covariance function.  Integrating with our problem, we denote the mean
function as $m(\mathbf{s})$ and the covariance function as
$k(\mathbf{s},\mathbf{s}^\prime)$, so a Gaussian process is
represented as:

{\footnotesize
\begin{equation}
\mathbf{\zeta}(\mathbf{s})  \sim
\mathcal{GP}_{j}(m(\mathbf{s}),k(\mathbf{s},\mathbf{s}^\prime)),\label{eq:gp}
\end{equation}
\par}
where

{\footnotesize
\begin{eqnarray}
m(\mathbf{s}) & = & E[\mathbf{\zeta}(\mathbf{s})], \nonumber\\
k(\mathbf{s},\mathbf{s}^\prime)& = & E[(\mathbf{\zeta}(\mathbf{s}) -
  m(\mathbf{s}))
  (\mathbf{\zeta}(\mathbf{s}^\prime) -m(\mathbf{s}^\prime))],\label{eq:mk_gp}
\end{eqnarray}
\par}

\subsubsection{Hyperparameter Optimization and Referencing}

We assume prediction noise as a Gaussian distribution and formulate
finding the optimal hyperparameters as an optimization problem.  We
seek the optimal solution of hyperparameters by maximizing the log
marginal likelihood (see~\cite{Carl06Gaussian} for details):

{\footnotesize
\begin{equation}
\log{p(\varPsi^{\prime}|\mathbf{s},\theta)} =
-\frac{1}{2}{\varPsi^{\prime}}^{T}K_{\varPsi^{\prime}}^{-1}\varPsi^{\prime} -
\frac{1}{2}\log{|K_{\varPsi^{\prime}}|}-\frac{n}{2}\log{2\pi},
\label{eq:loglikelihood}
\end{equation}
\par}
where $K_{\varPsi^{\prime}}$ is the calculated covariance matrix of
the target vector. 

With the optimal hyperparameters, the prediction distribution is represented as:

{\footnotesize
\begin{eqnarray}
{\varPsi^{\prime}}^{\ast}|\mathbf{s}^{\ast},\mathbf{s},\varPsi^{\prime}
 \sim  \mathcal{N}(\mathbf{k}(s^{\ast}, \mathbf{s})^{T}[K +
  \sigma_{noise}^{2}I]^{-1}\varPsi^{\prime},  \nonumber \\
k(s^{\ast},s^{\ast})+\sigma_{noise}^{2}-\mathbf{k}(s^{\ast},\mathbf{s})^{T}[K
  + \sigma_{noise}^{2}I]^{-1}\mathbf{k}(s^{\ast},\mathbf{s})), \label{eq:GPRreference}
\end{eqnarray}
\par}
where $K$ is the calculated covariance matrix from training 2D image
features $\mathbf{s}$ and $\sigma_{noise}$ is the covariance of
Gaussian noise.

Equation\ref{eq:GPRreference} for referencing test data is deducted from marginal and conditional properties of Gaussian distributions. The following is the marginal property of Gaussian distributions: the marginal of a joint Gaussian is again a Gaussian, that is,

{\footnotesize
\begin{eqnarray}
\displaystyle && p(\mathbf{x},\mathbf{y}) = \mathcal{N}( \left[ \begin{array}{c}
a  \\
b
 \end{array} \right],\left[ \begin{array}{cc}
A & B  \\
B^{T} & C
 \end{array} \right])\quad  \nonumber\\
&\Rightarrow& \quad  p(\mathbf{x}) =
\mathcal{N}(\mathbf{a},A). \end{eqnarray}
\par}
And the conditional property of Gaussian distributions are: the conditionals of a joint Gaussian are again Gaussian, that is,

{\footnotesize
\begin{eqnarray}
\displaystyle p(\mathbf{x},\mathbf{y})&=& \mathcal{N}(\left[ \begin{array}{cc}
a  \\
b
 \end{array} \right],\left[ \begin{array}{cc}
A & B  \\
B^{T} & C
 \end{array} \right])\quad  \nonumber\\
\Rightarrow \quad
p(\mathbf{x}|\mathbf{y}) &=&
\mathcal{N}(\mathbf{a}+BC^{-1}(\mathbf{y}-\mathbf{b}),
A-BC^{-1}B^{T}). \end{eqnarray}
\par}
Thus we are able to predict the distribution of $x$ given the distribution $y$.

In most cases, we assume that Gaussian process priors have zero means, that is,

{\footnotesize
\begin{equation} f(x)|M_{i} \sim \mathcal{GP}_{j}(m(\mathbf{x}) \equiv
0,k(\mathbf{s},\mathbf{s}^\prime)).
\end{equation}
\par}
This leads to a Gaussian process posterior

{\footnotesize
\begin{equation}
\qquad \,  f(x)|\mathbf{x},\mathbf{y}, M_{i} \sim
\mathcal{GP}_{j}(m_{post}(\mathbf{x}), k_{post}(\mathbf{s},\mathbf{s}^\prime)),
\end{equation}
\par}
where

{\footnotesize
\begin{equation}
m_{post}(\mathbf{x}) = k(x, \mathbf{x})[K(x,\mathbf{x}) + \sigma_{noise}^{2}I]^{-1}\mathbf{y}.
\end{equation}
\par}
With this posterior, we only need to define covariance matrices, known as kernel in machine learning community.

The most frequently used covariance matrices (kernels) include: squared exponential (SE), Rational quadratic (RQ), Mat\'{e}rn and Periodic, smooth covariance functions. The function of covariance function is to define the distance measure in a newly transformed space where the original data samples have one to one correspondences with their mapped points and due to the transformation, data samples of different attribute classes in the new spaces are easier to classify or identify. With the kernel trick, we can get rid of directly defining the mapping model and only define the kernel matrix, the covariance matrix here.

\subsubsection{GPR for 2D to 3D Pose Mapping}

Gaussian processes yield a method for specifying a probability distribution over functions by specifying a mean and a covariance function for the function values $f(x)$. By training a Gaussian process with sample data $\{x, f(x)\}$ the variance of the Gaussian process becomes small for function values $f(x)$ at supporting points $x$ included in
the training data, which corresponds to an increased certainty about the function values at these points, while at other points $x'$ the variance of the Gaussian process remains
high which corresponds to a high uncertainty about the function values $f(x')$ at such points.

In our algorithm, we select the most commonly used covariance matrix: squared exponential covariance matrix. Given a 2D pose estimate which is represented as the 26 dimensional vector $BP$ ($13*2$, where $13$ is the number of body joints in MoP and $2$ is the dimension size), we train one Gaussian process
to predict each of the 60 dimensions of the 3D pose vector $\psi$ ($20*3$, where $20$ is the number of body joints in HumanEva motion capture data and $3$ is the dimension size) separately. Then given features from test samples, we predict 3D poses with trained GPR.

%
%
%
%
%
%
%
%
%
%
%

\section{Results}\label{sec:results}

To demonstrate the effectiveness of the proposed method, we first show improved 2D body part localization results with the proposed feature fusion method; then, the trained 2D to 3D pose estimator is carried out on detected 2D body part locations and 3D poses are estimated and shown.

\subsection{Evaluation Data and Experiment Settings}
From HumanEva-I data set, we select two different actions (``Walking'' and ``Box'') performed by three different actors (``S1'', ``S2'' and ``S3''). All three performers perform the actions within a fixed area (confined with a carpet).  ``Walking'' is performed in a cyclic way, while in ``Box'', performers are moving in a very small area positions notwithstanding different performing style.

As a result, we have six different experiments in total. Training and test are carried out within each experiment. That is, we train a detector on a single experiment setting and validate the trained models on the test frames of same experiment setting. This experiment setting is designed to compare the influence of different action type and different performer to the body part localization and pose estimation results. The detailed splits between training and test is shown in the following table.
\begin{table}[t!]
\begin{center}
\begin{tabular} {|c|c|c|c|c|}
\hline  {Exp.} & {Action} & { Actor }  &  {TrFrmNo} & {TeFrmNo}\\
\hline
\hline  {1}& {Walking} & {$S1$} &  {$200$} &  {$21$}\\
\hline  {2}& {Walking} & {$S2$} &  {$200$} &  {$21$}\\
\hline  {3}& {Walking} & {$S3$} &  {$200$} &  {$21$}\\
\hline  {4} & {Box} &  {$S1$}    & {$200$} &  { $21$}\\
\hline  {5} & {Box} &  {$S2$}    & {$200$} &  { $21$}\\
\hline  {6} & {Box} &  {$S3$}    & {$200$} &  { $21$}\\
\hline
\end{tabular}
\end{center}
\caption{The composition of experiments from Humaneva data set. Exp. ``TrFrmNo'' denotes total frame numbers for training. ``TeFrmNo'' denotes total frame numbers for test.}
 \label{table:trainingComp}
\end{table}

For each action performed by a specific actor, training data are composed of $200$ consecutive frames, which is close to the number of frames in a cyclic walking sequence. Test data are sampled with a equal step among the whole motion sequence excluding the training frames, so that the test poses covers all possible poses for an action.

\subsection{Enhanced 2D Part Locations}

The proposed 2D part localization method aims to solve double counting problem, that is, a pixel location (a body part location in our case) might be designated to two body limb positions even there is no occlusions between these body limbs. The reason for the double counting problem in MoP detection is:
\begin{enumerate}
\item the responses of a pixel location (or a body part observation) to all trained body part templates are calculated separately,
\item then, from leaf nodes to root nodes, a best response is selected for a each node among all candidate mixtures and this response is passed as a message to its parent node, that is, a locally optimal solution.
\end{enumerate}
The limitation of this solution is that usually, the chained body part position calculated from local optimums are not a global optimum and the essence of the solution gives no globally target.
\begin{figure*}[!ht]
\centering
\renewcommand{\tabularxcolumn}[1]{>{\arraybackslash}m{#1}}
\begin{tabularx}{\textwidth}{cccc}
    MoP detection & Our detection & MoP detection & Our detection \\\\
    \includegraphics[width=0.24\linewidth]{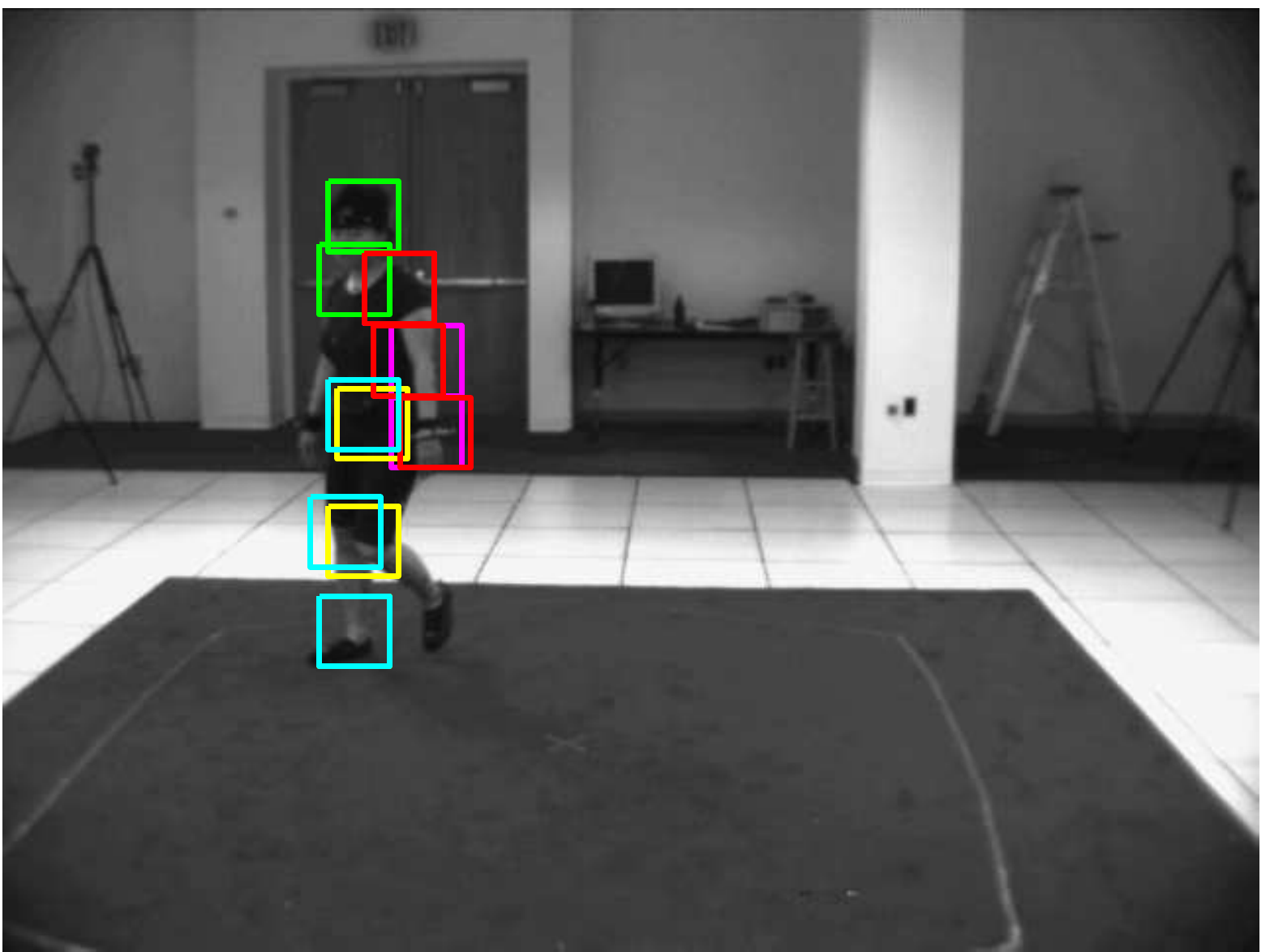} &
\includegraphics[width=0.24\linewidth]{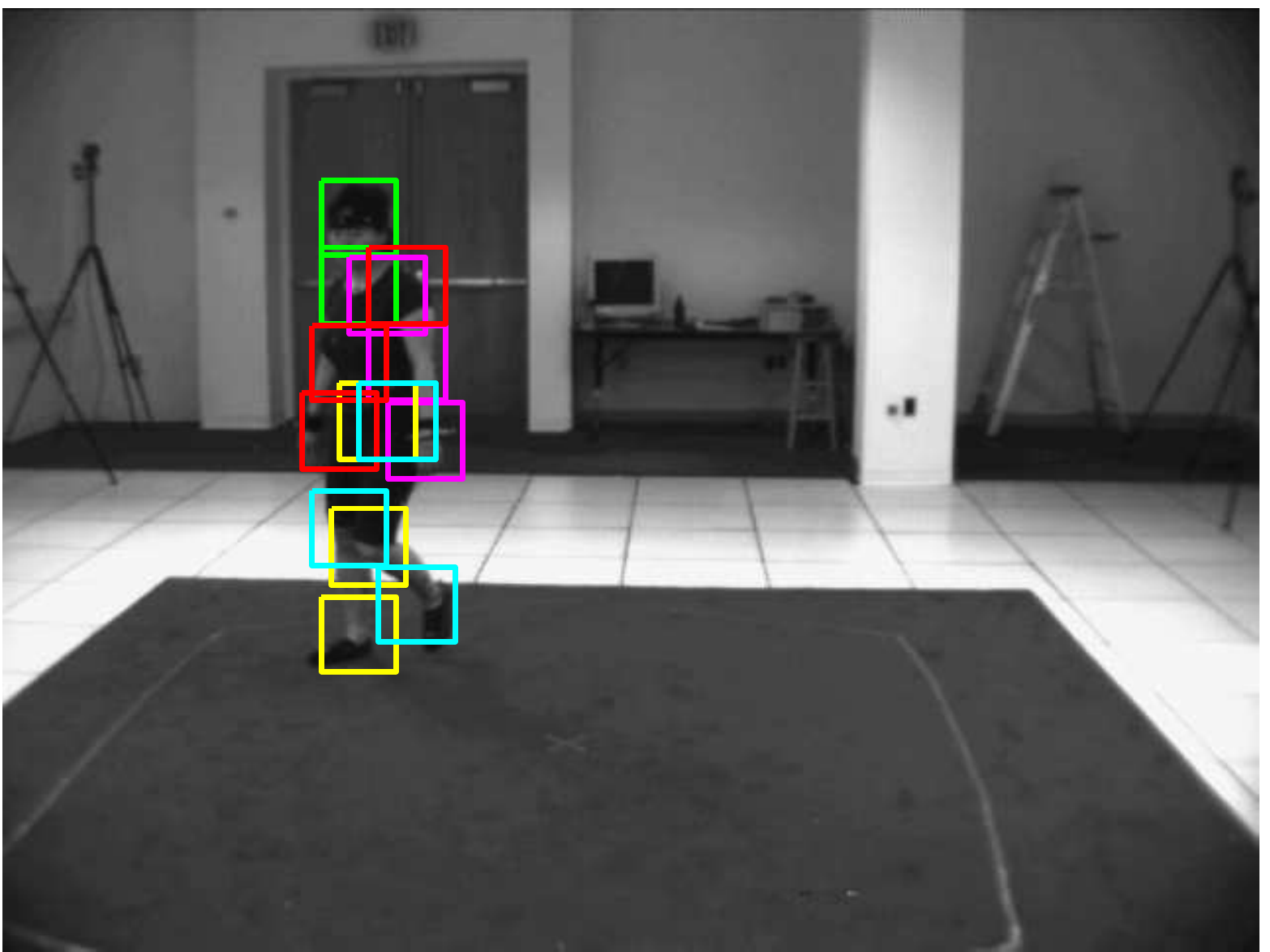}  & \includegraphics[width=0.24\linewidth]{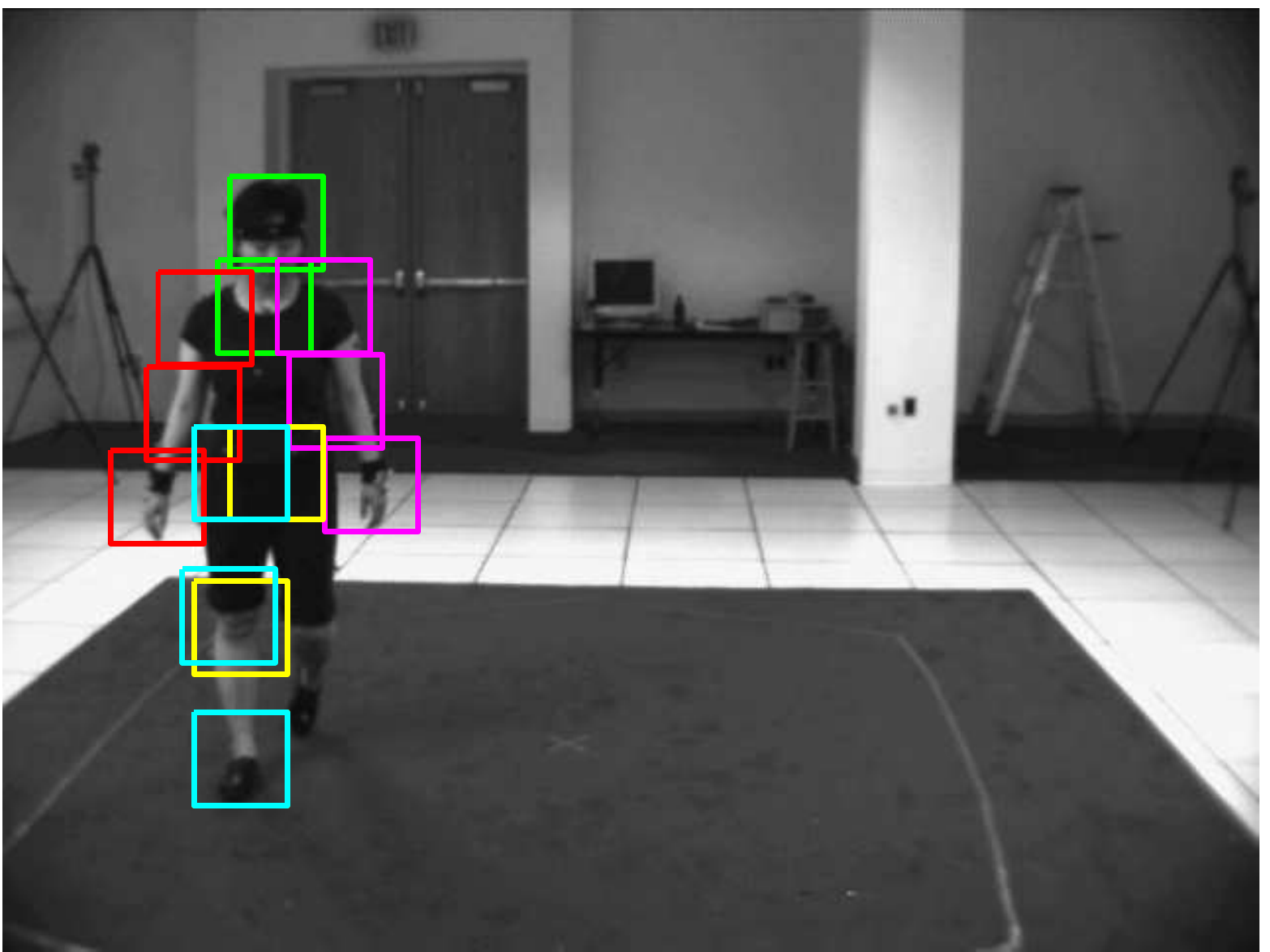} &
\includegraphics[width=0.24\linewidth]{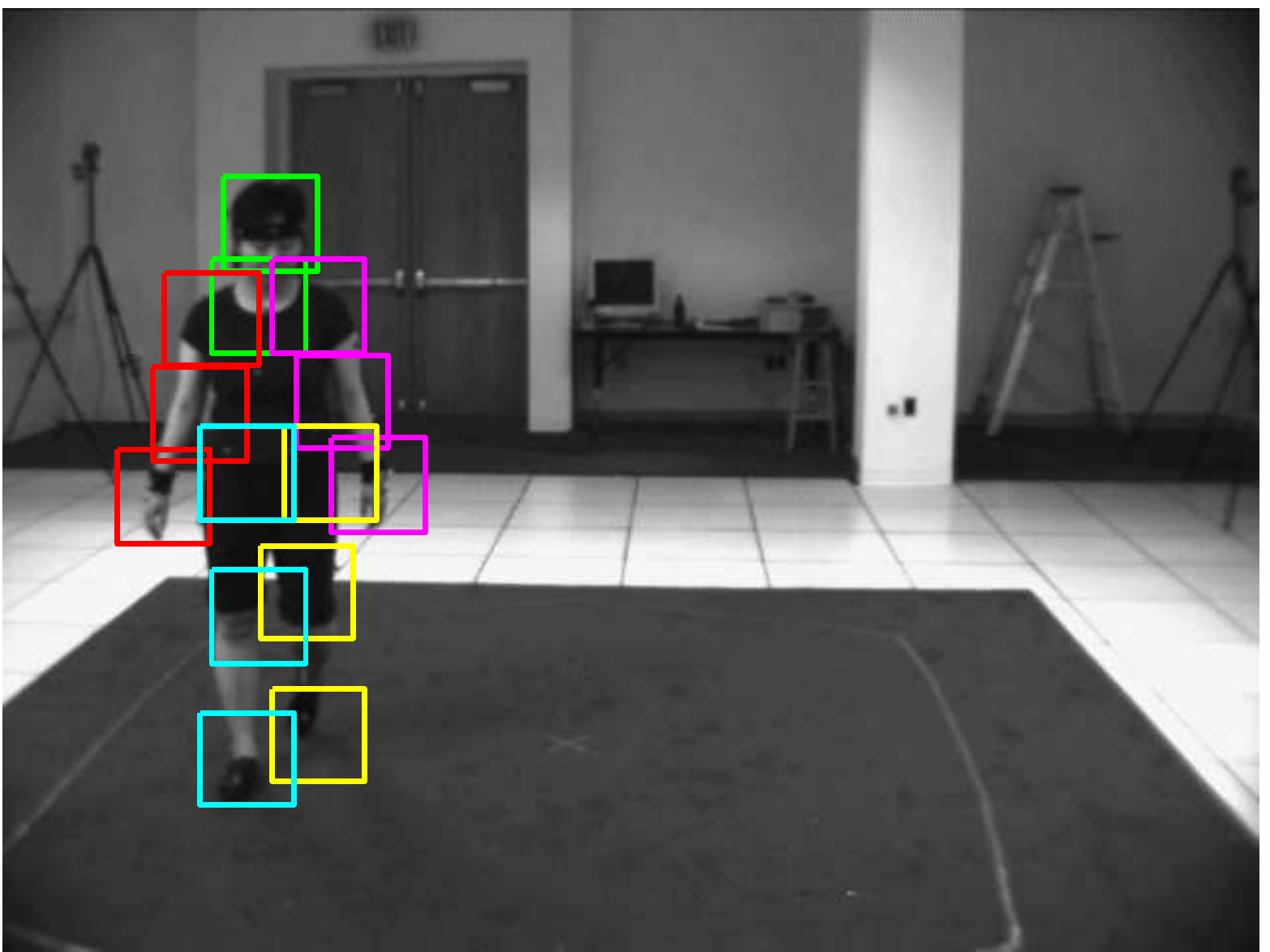} \\
  \includegraphics[width=0.24\linewidth]{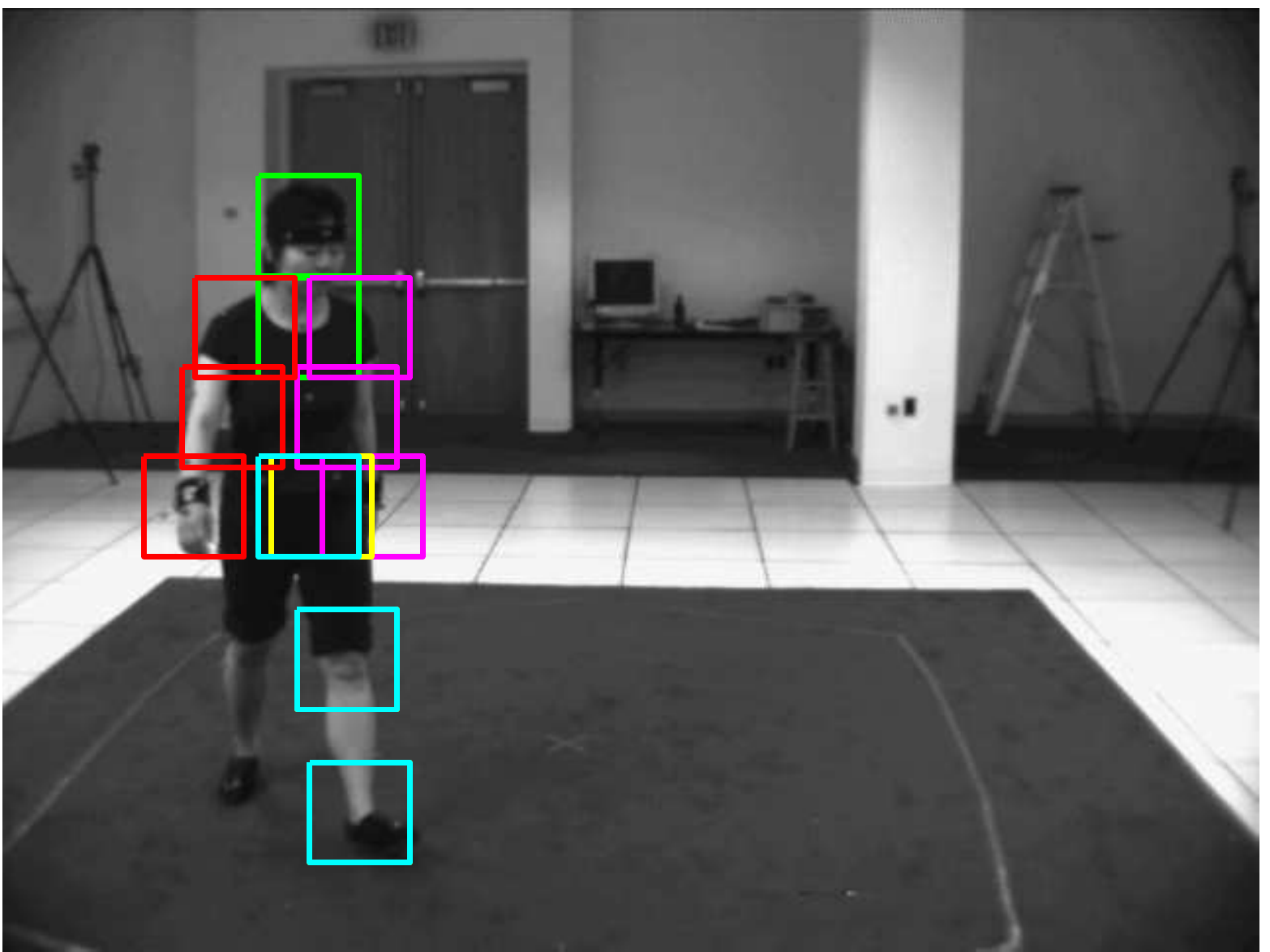} &
\includegraphics[width=0.24\linewidth]{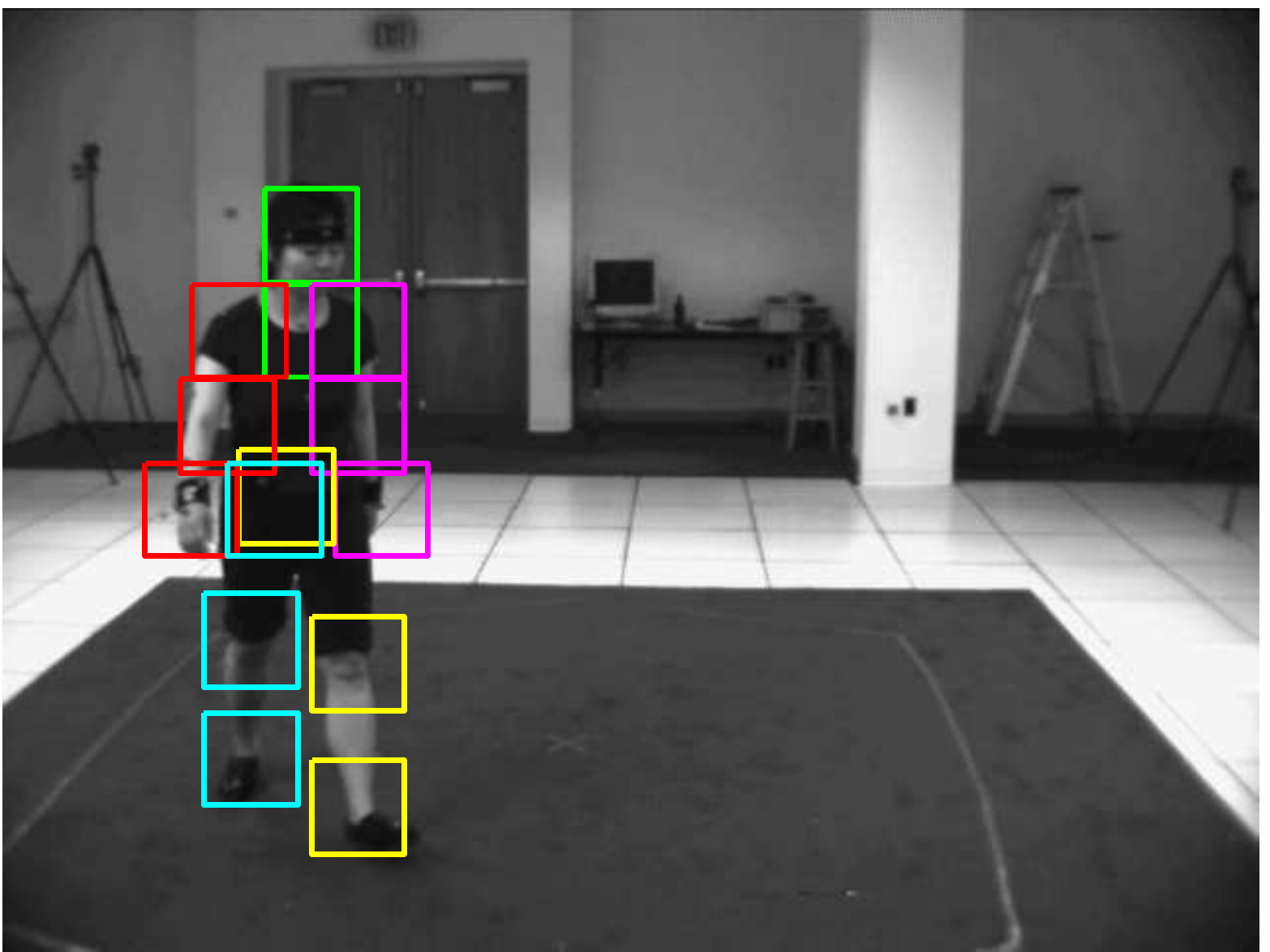} & \includegraphics[width=0.24\linewidth]{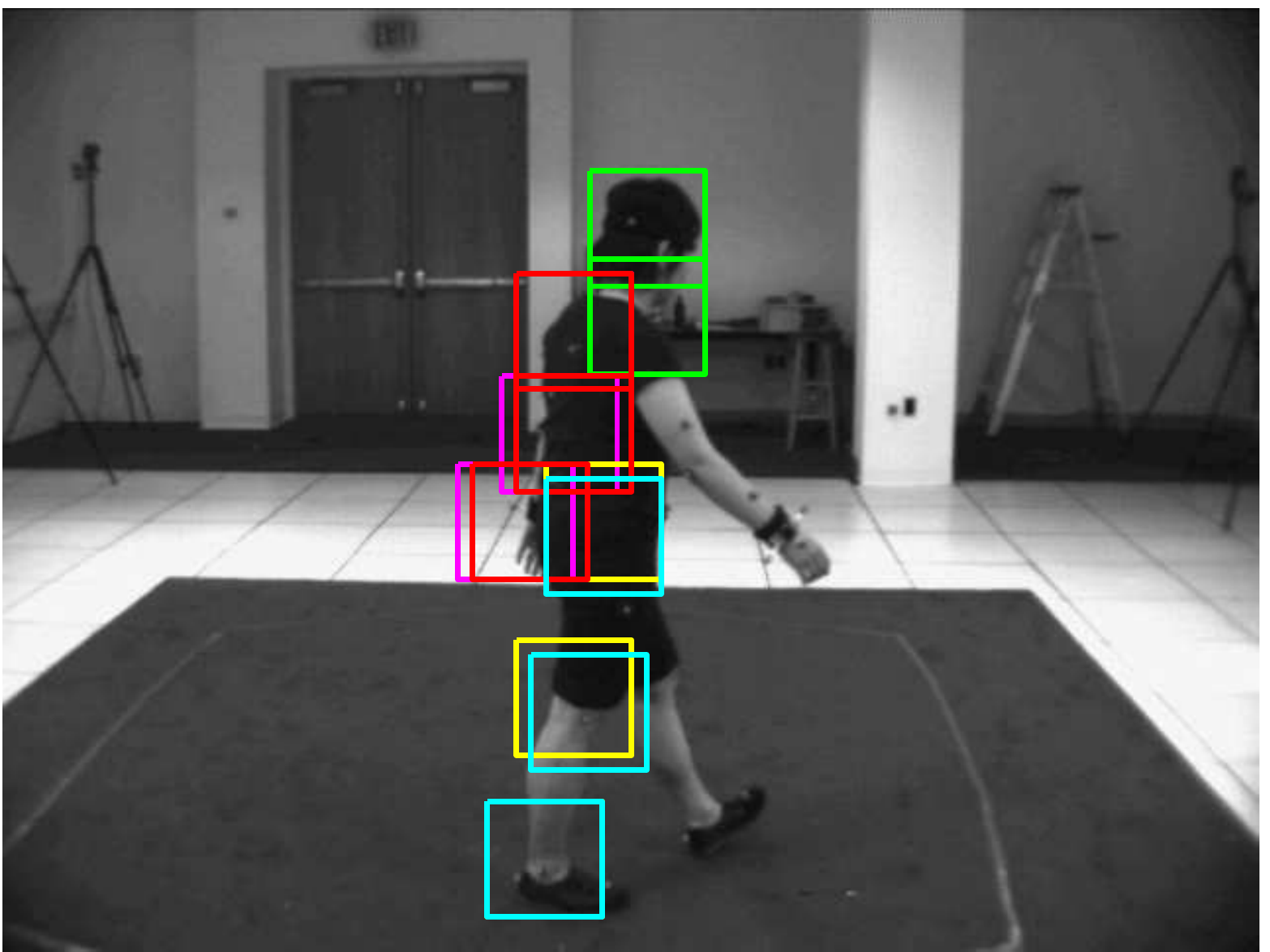} &
\includegraphics[width=0.24\linewidth]{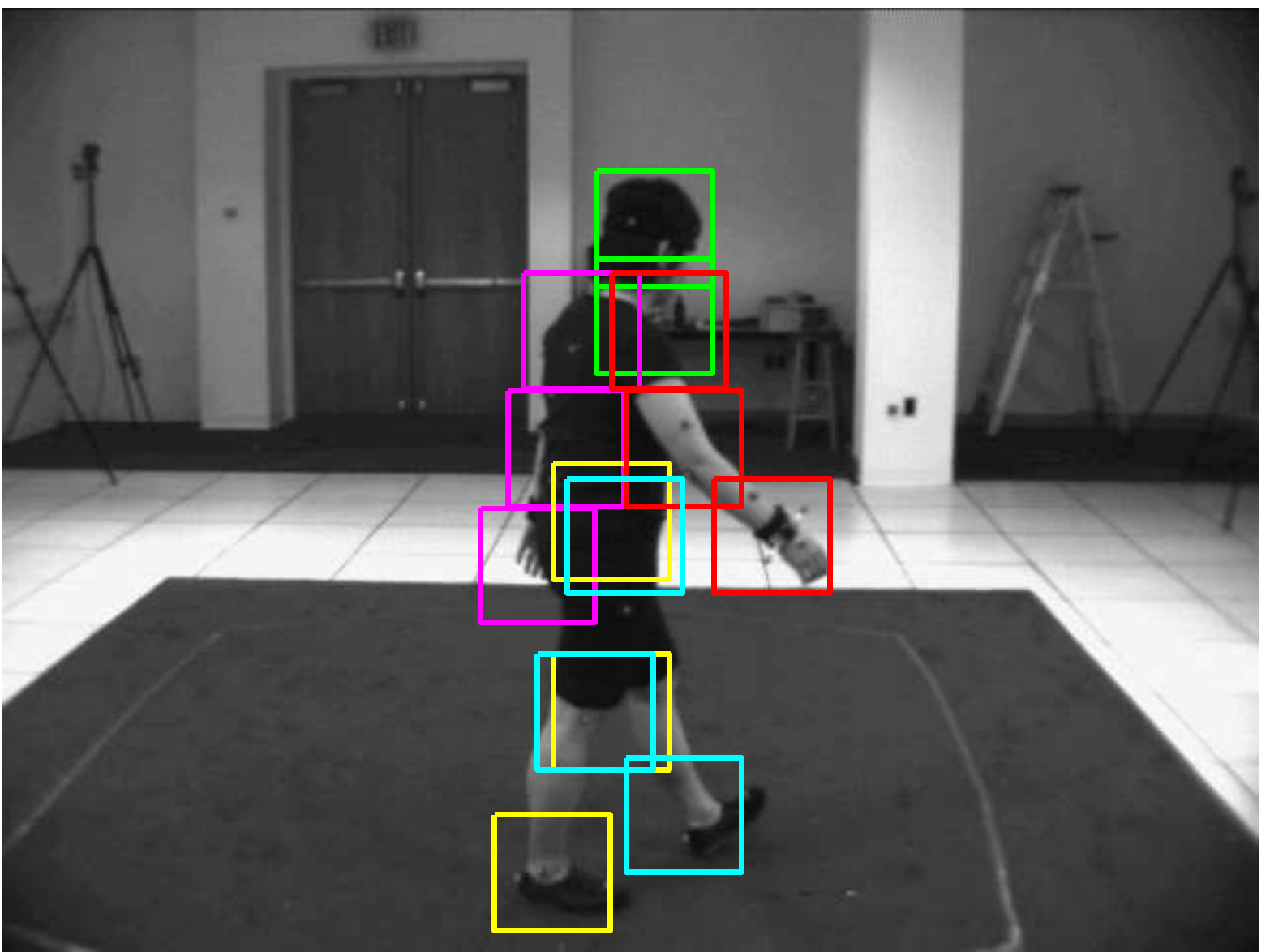} \\
   \includegraphics[width=0.24\linewidth]{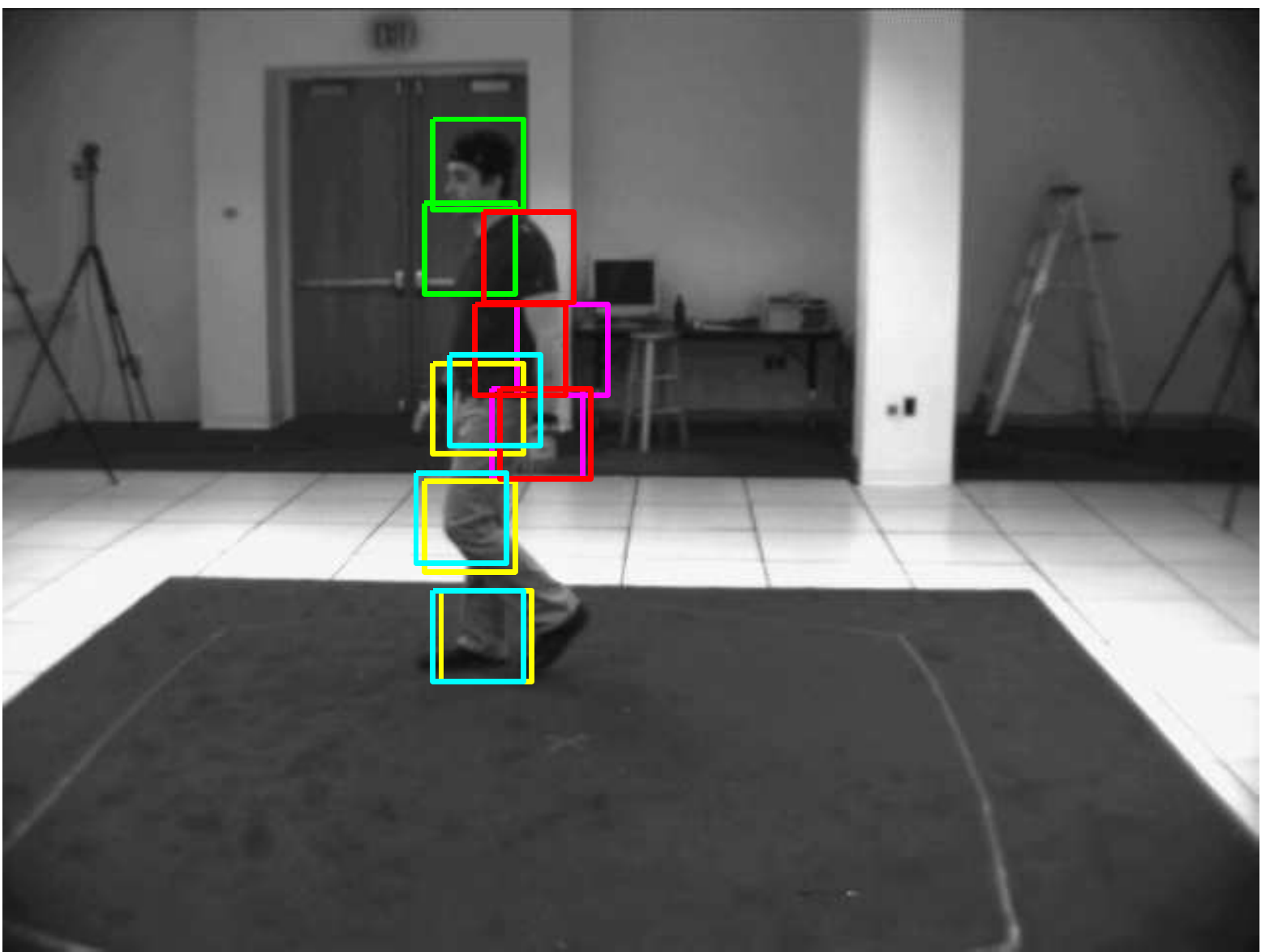} &
\includegraphics[width=0.24\linewidth]{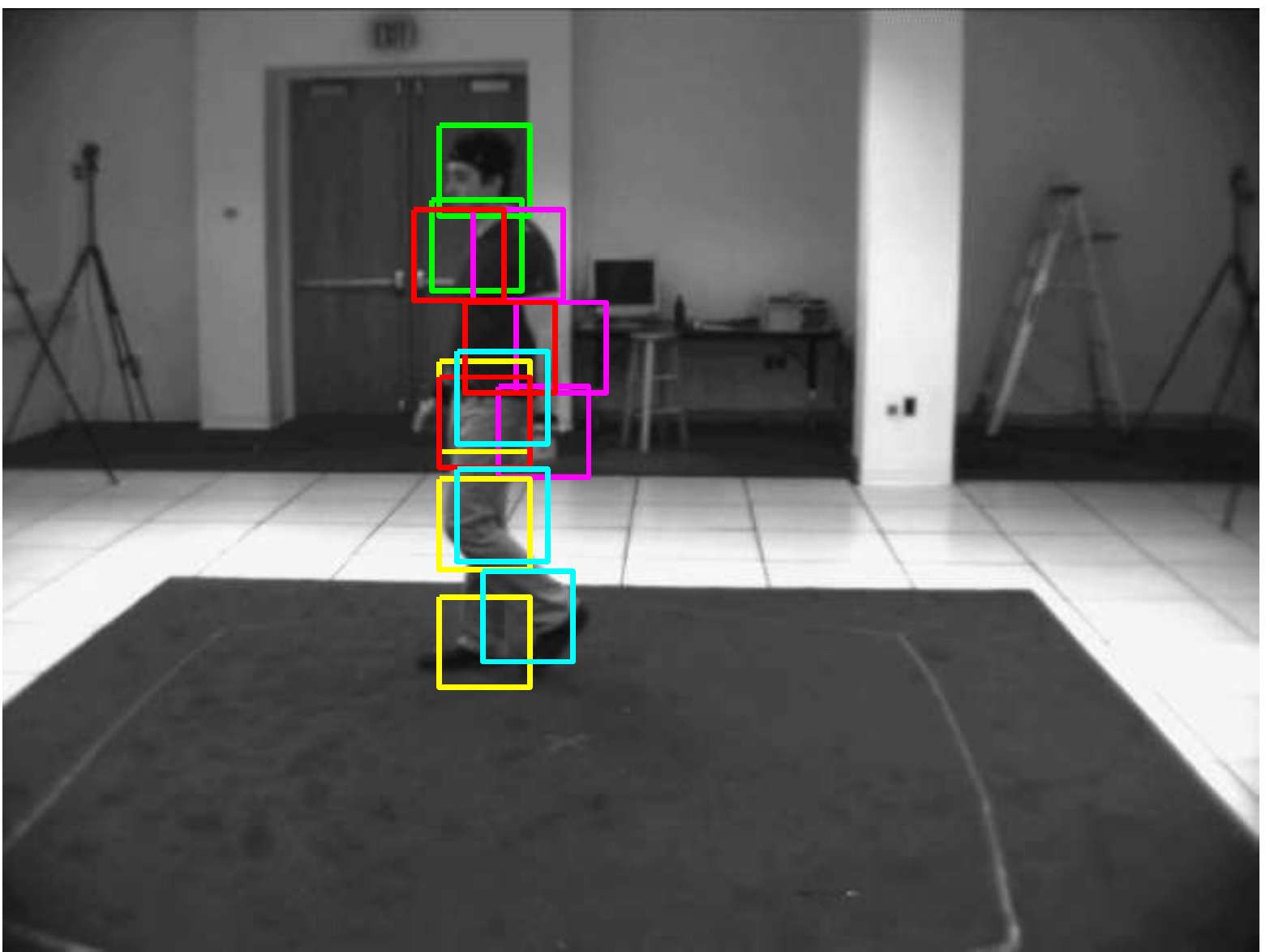} & \includegraphics[width=0.24\linewidth]{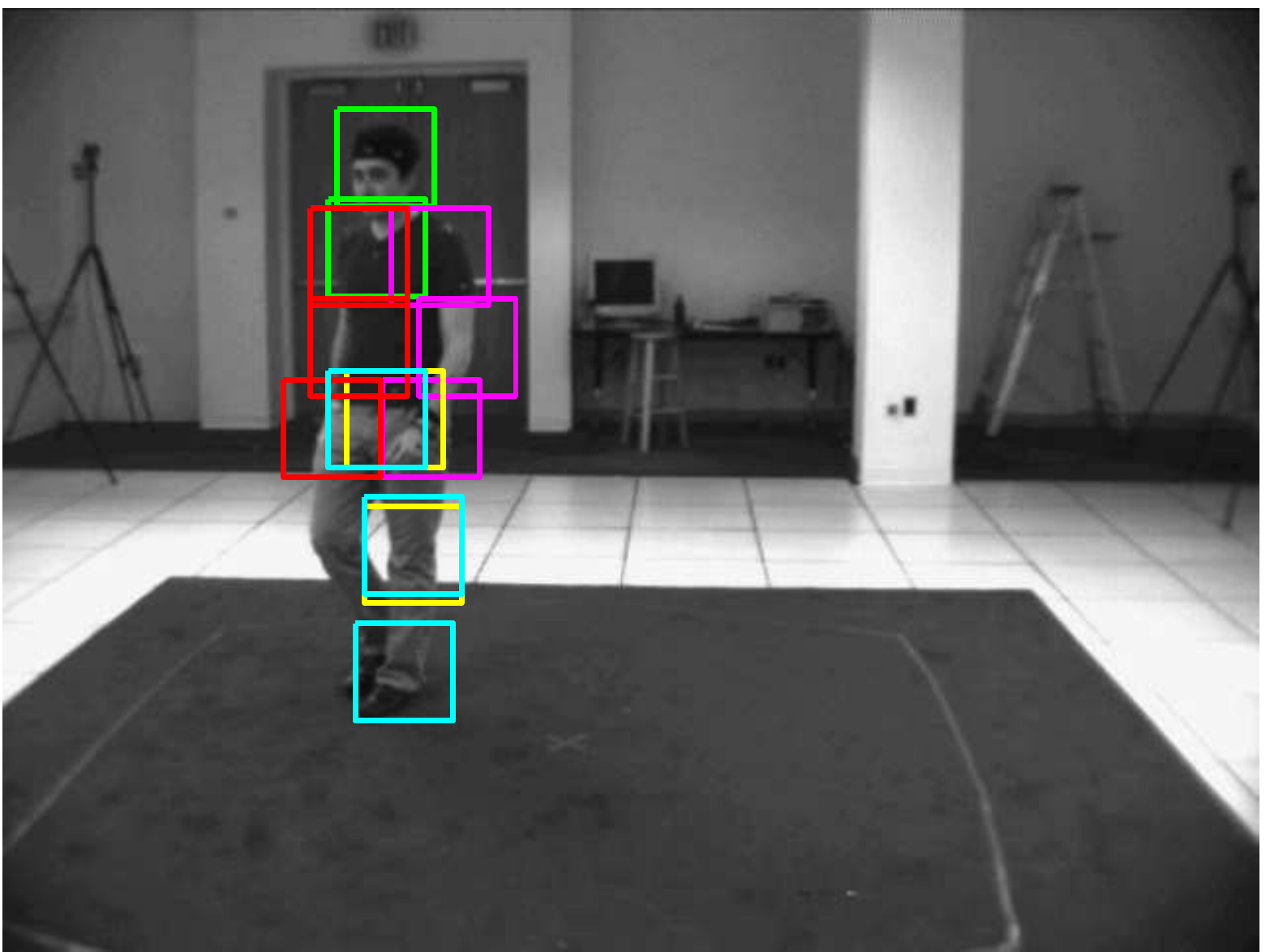} &
\includegraphics[width=0.24\linewidth]{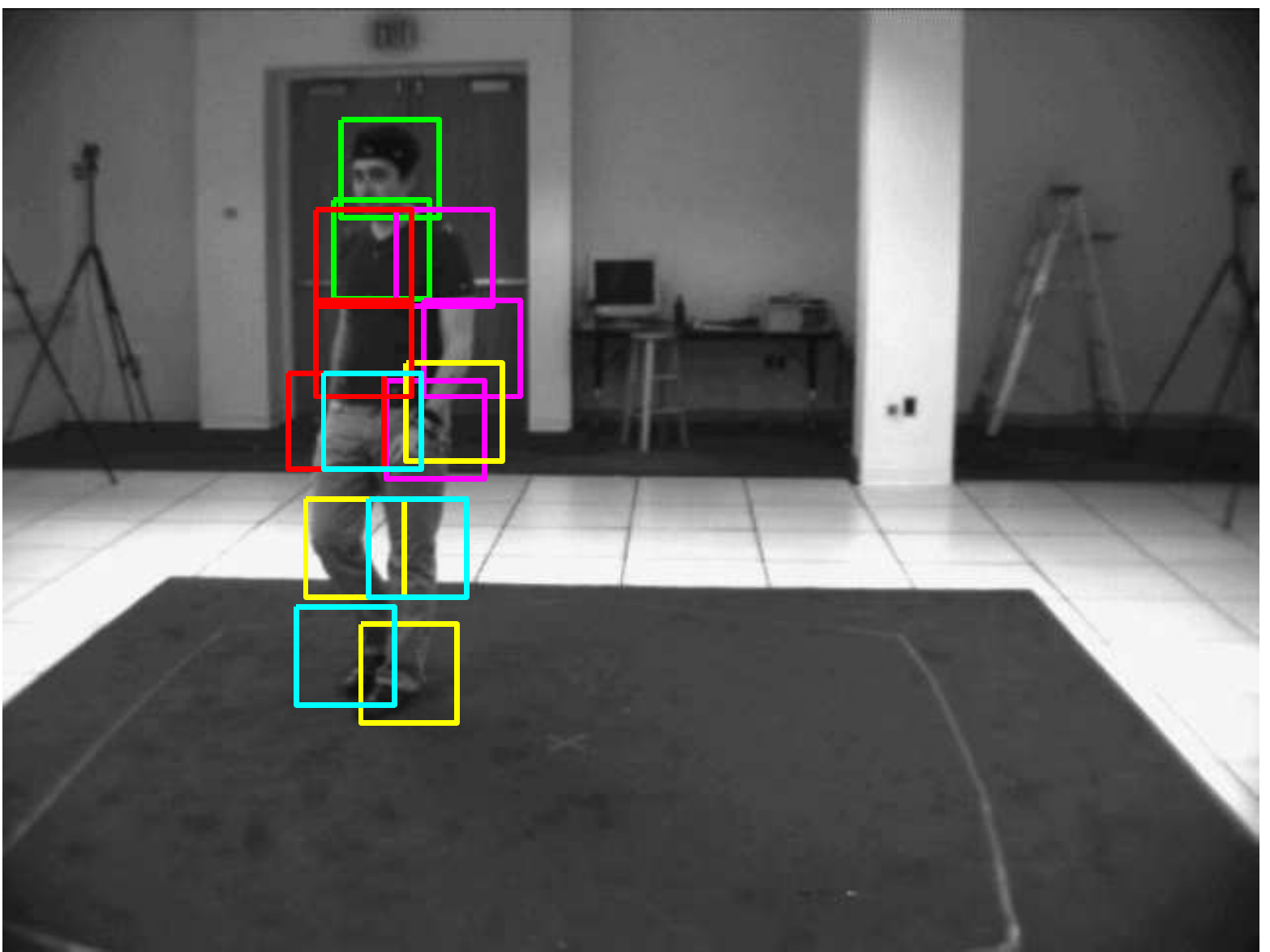} \\
    \includegraphics[width=0.24\linewidth]{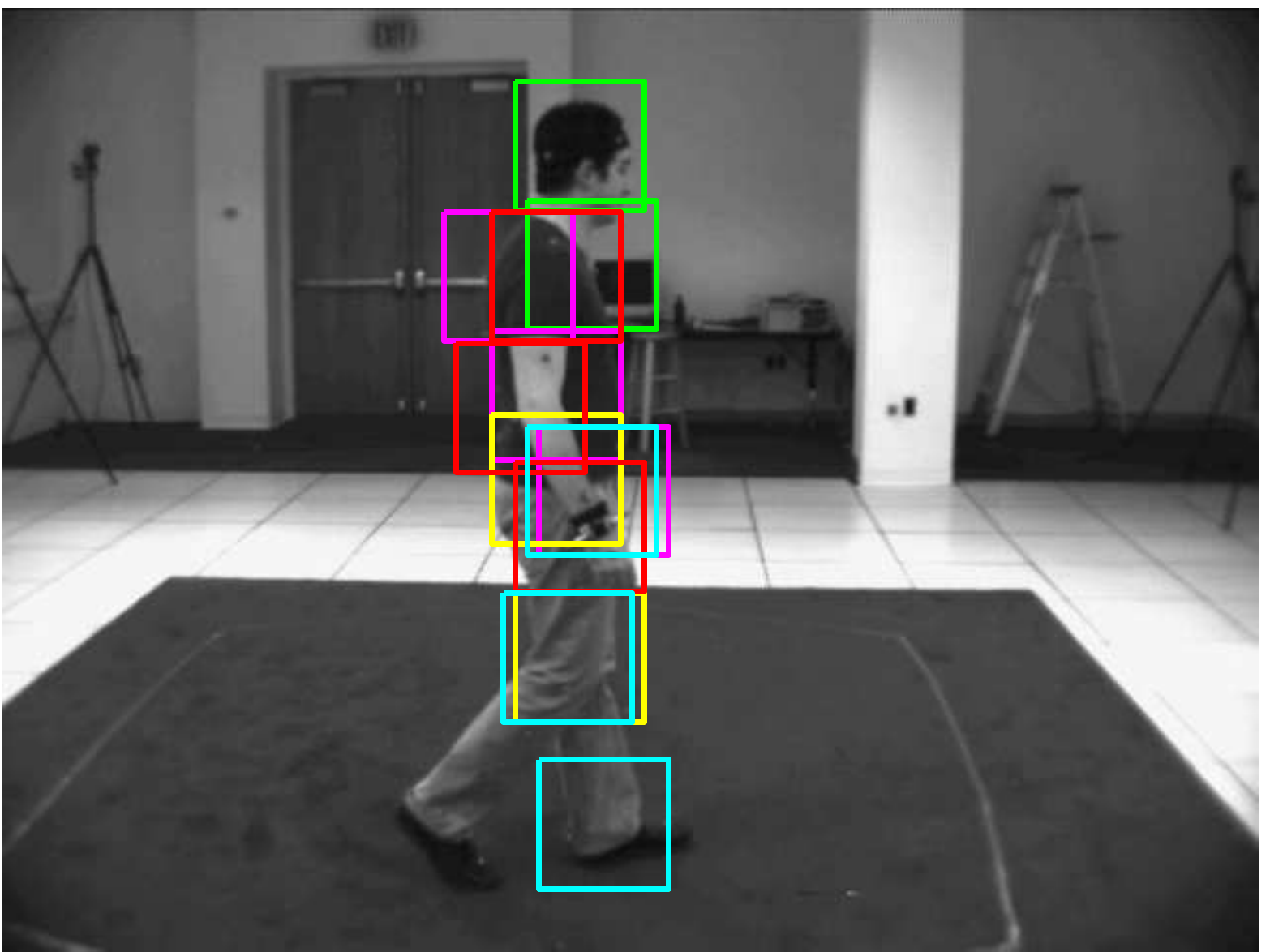} &
\includegraphics[width=0.24\linewidth]{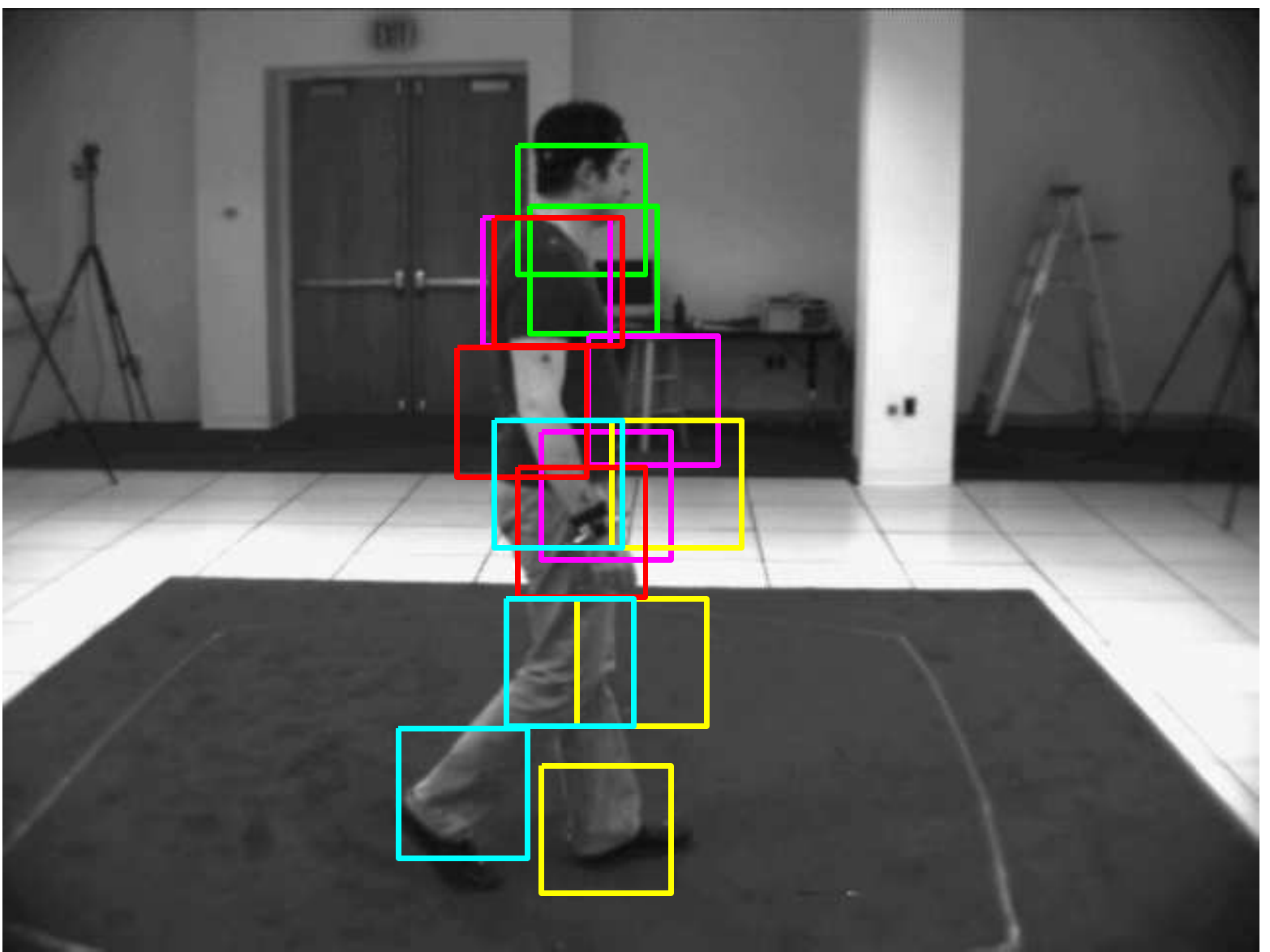} & \includegraphics[width=0.24\linewidth]{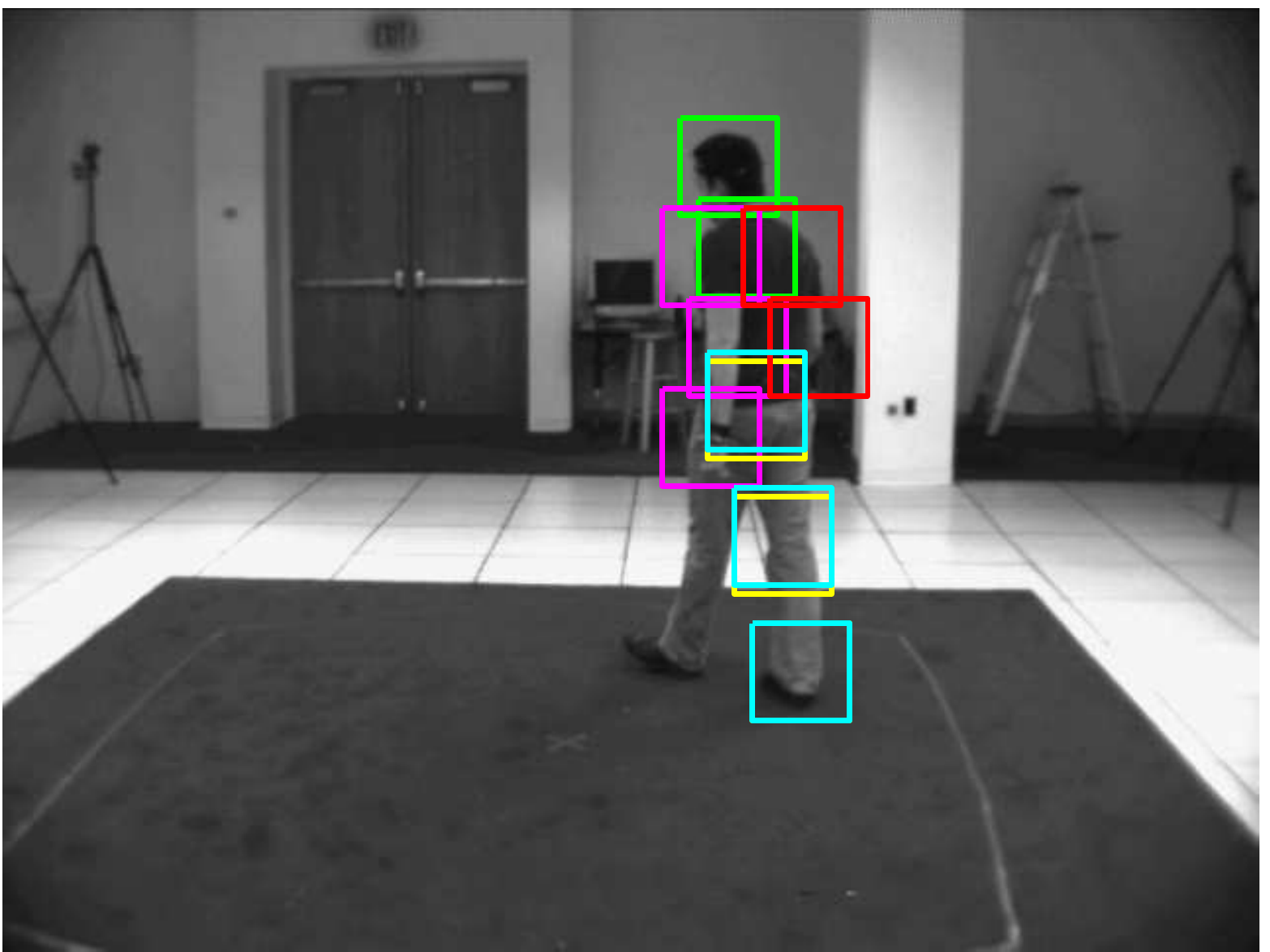} &
\includegraphics[width=0.24\linewidth]{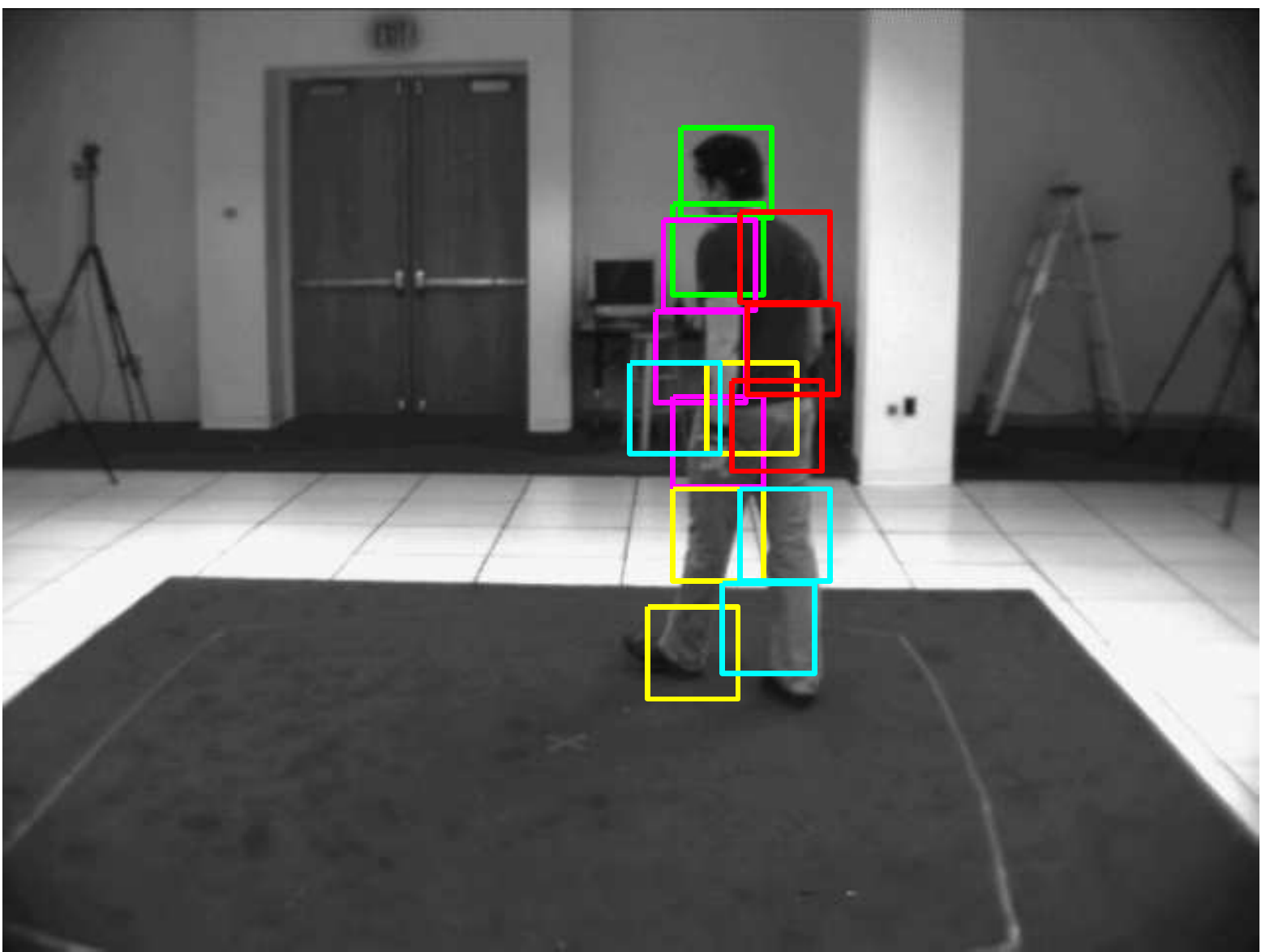}\\
    \includegraphics[width=0.24\linewidth]{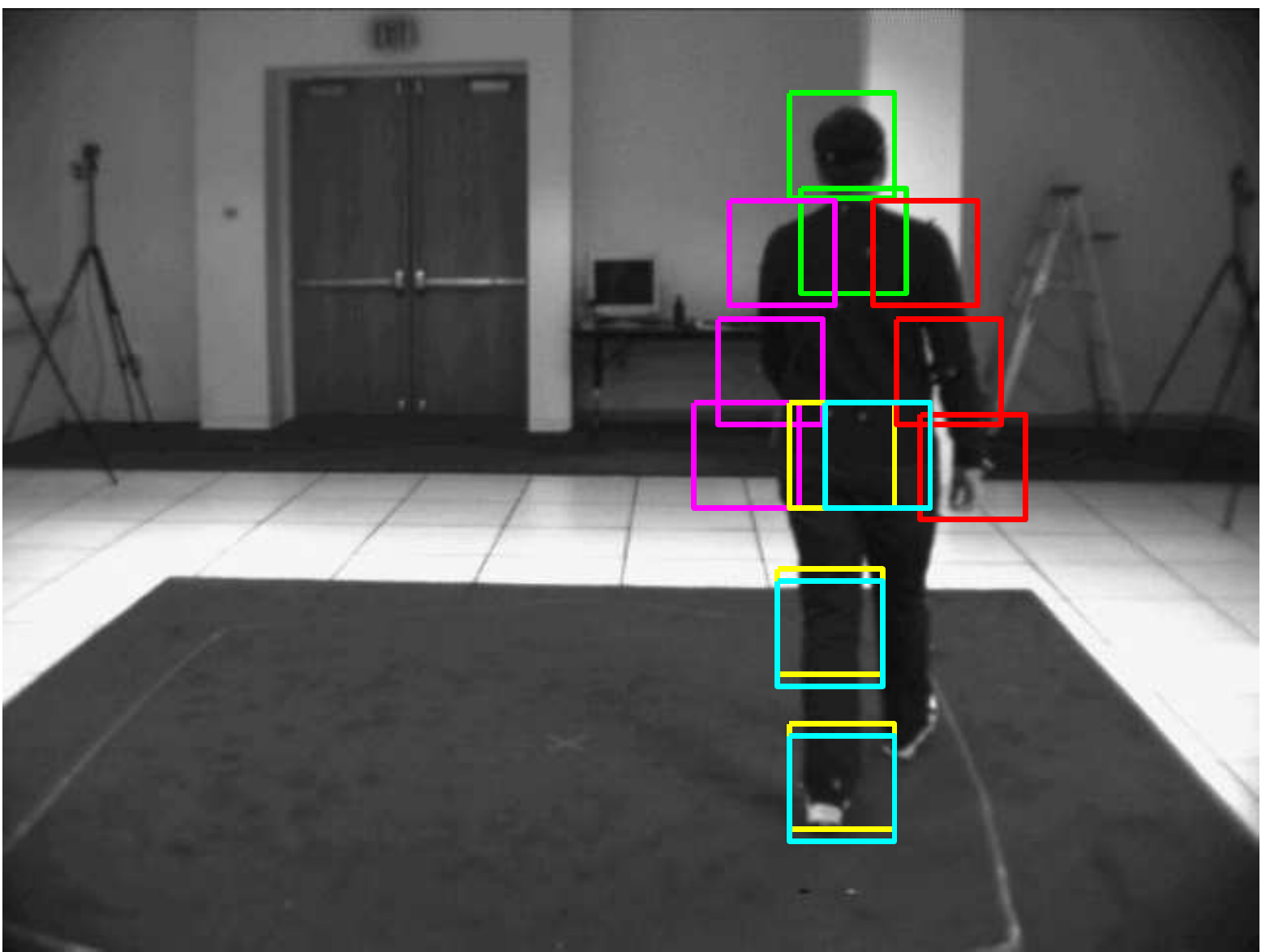} &
\includegraphics[width=0.24\linewidth]{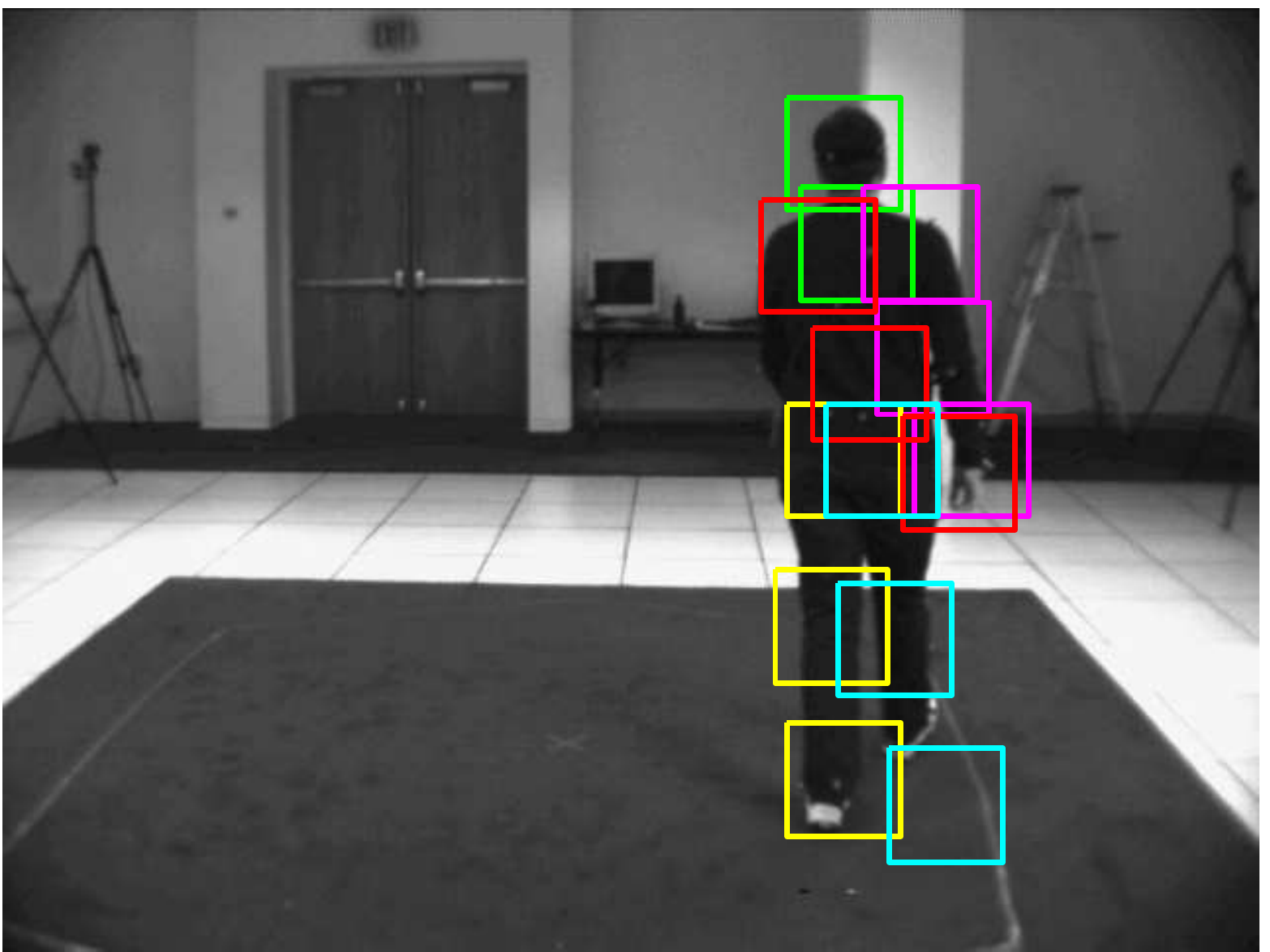} & \includegraphics[width=0.24\linewidth]{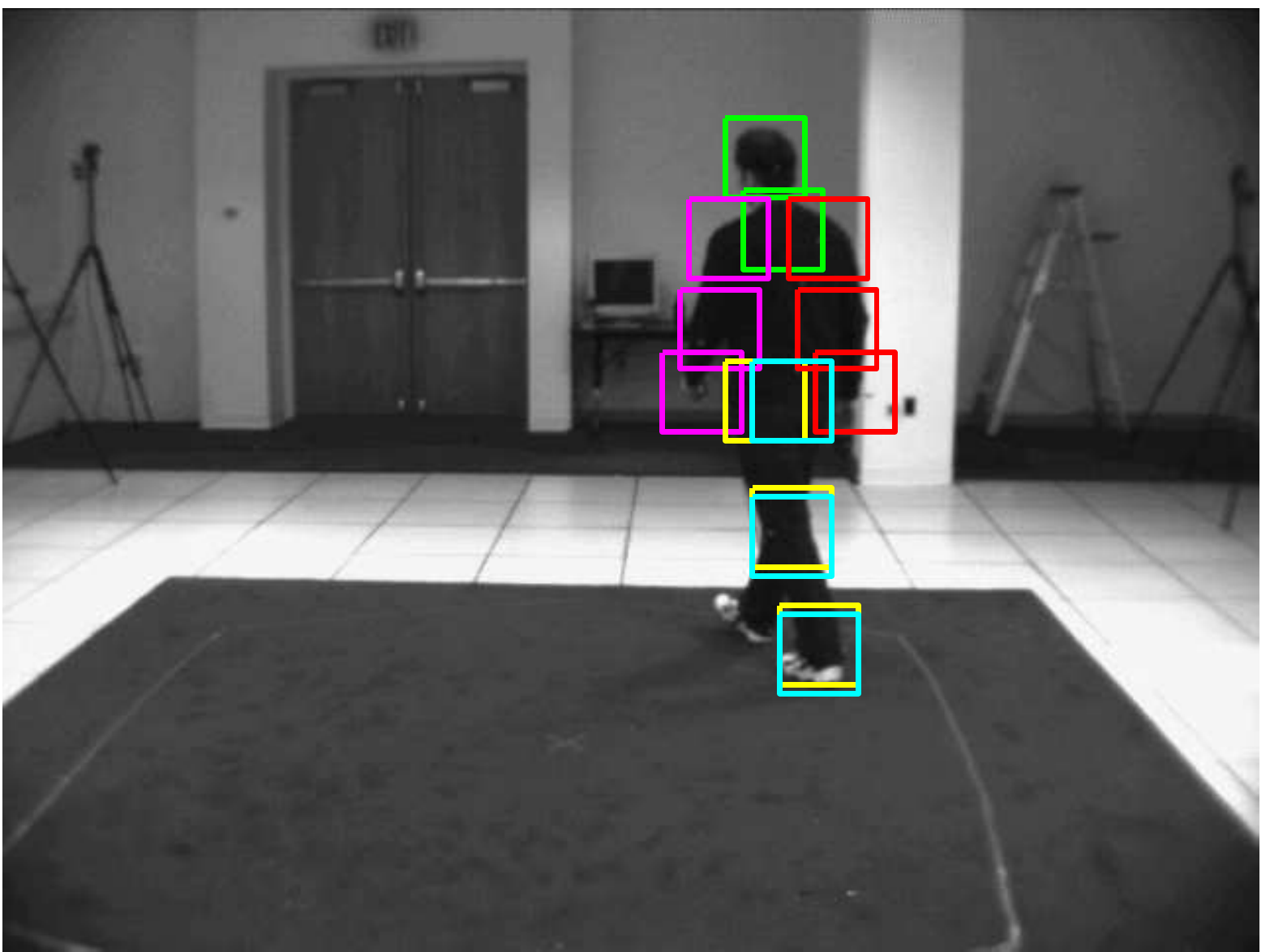} &
\includegraphics[width=0.24\linewidth]{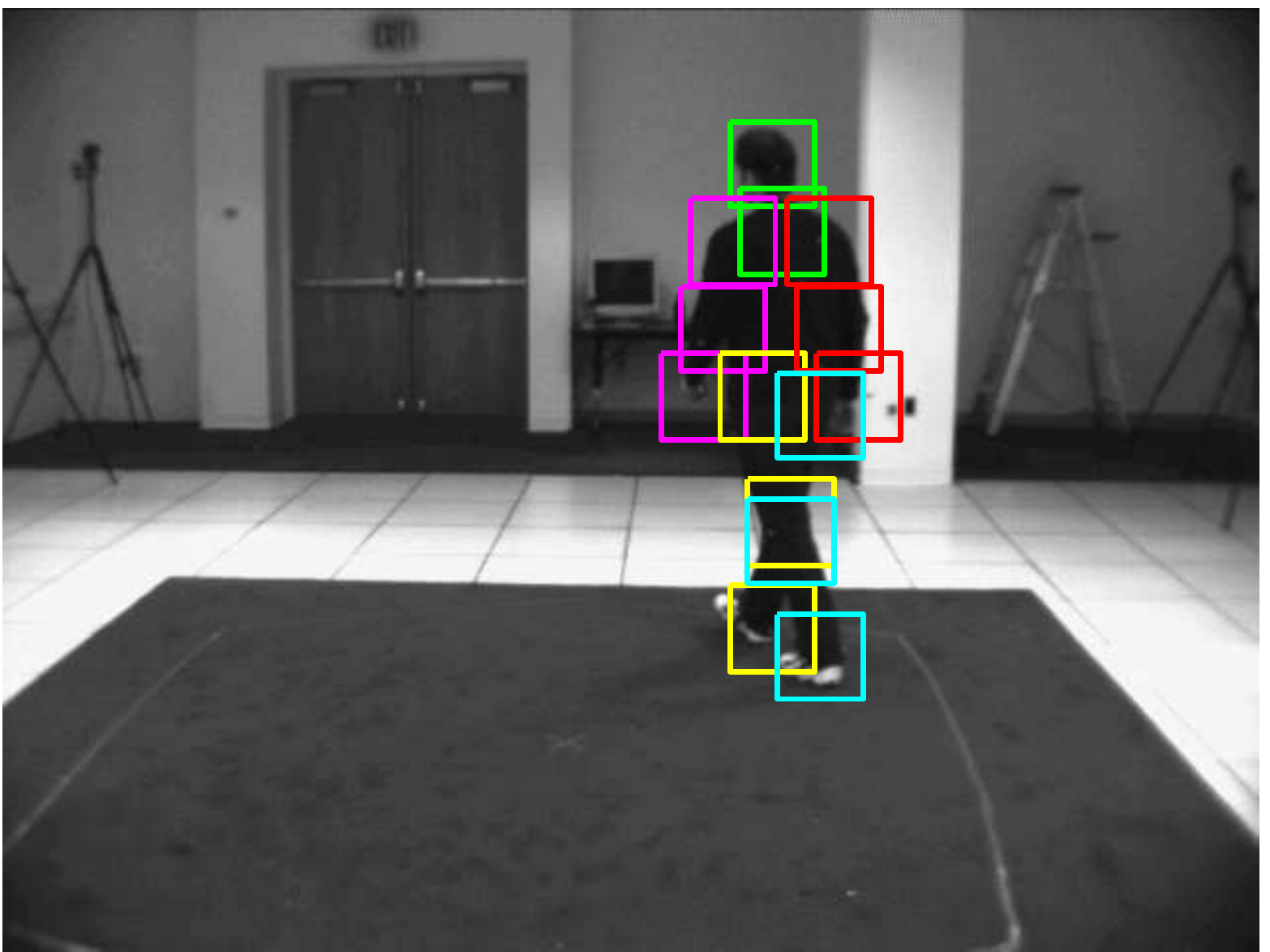}\\
    \includegraphics[width=0.24\linewidth]{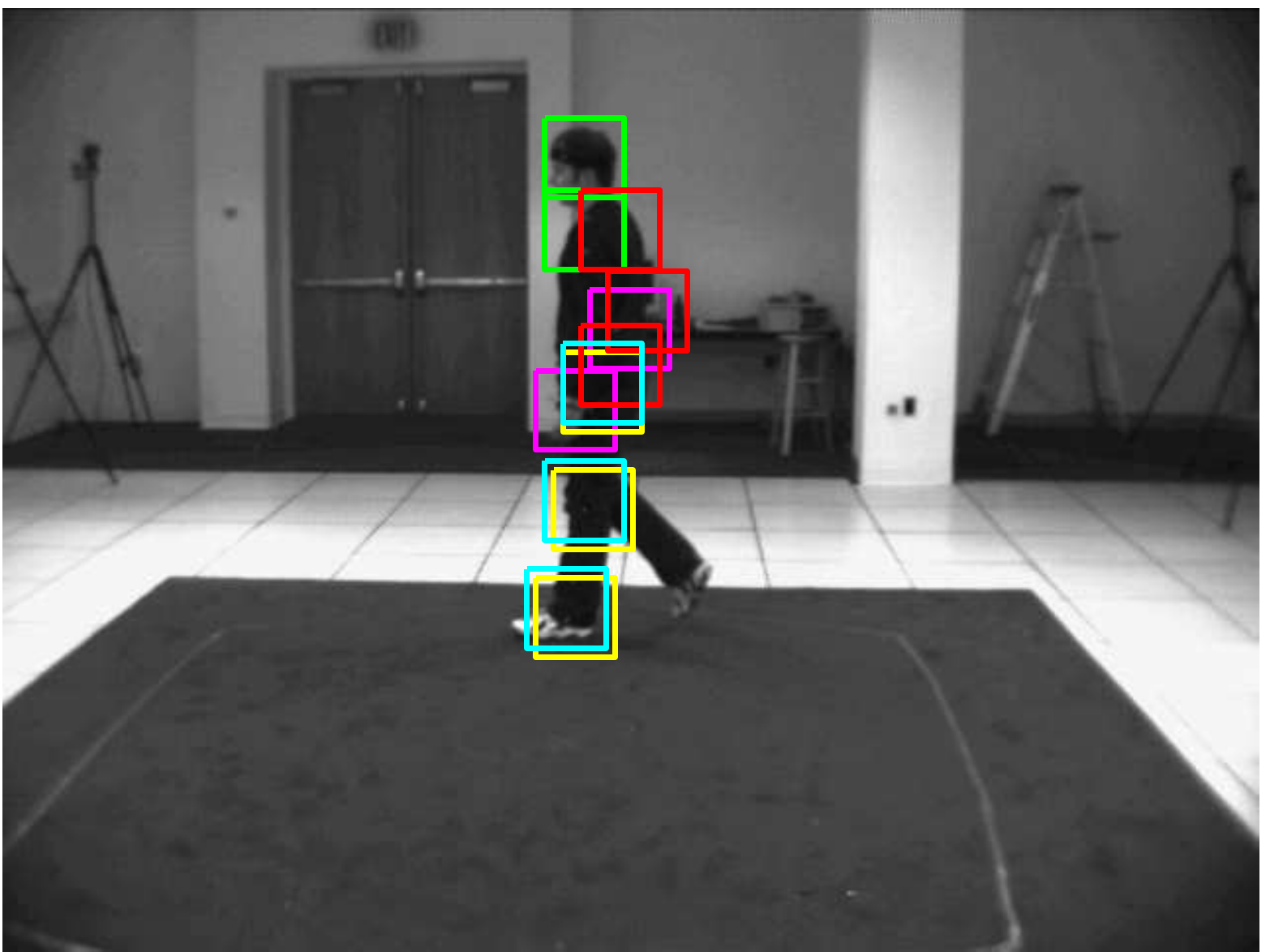} &
\includegraphics[width=0.24\linewidth]{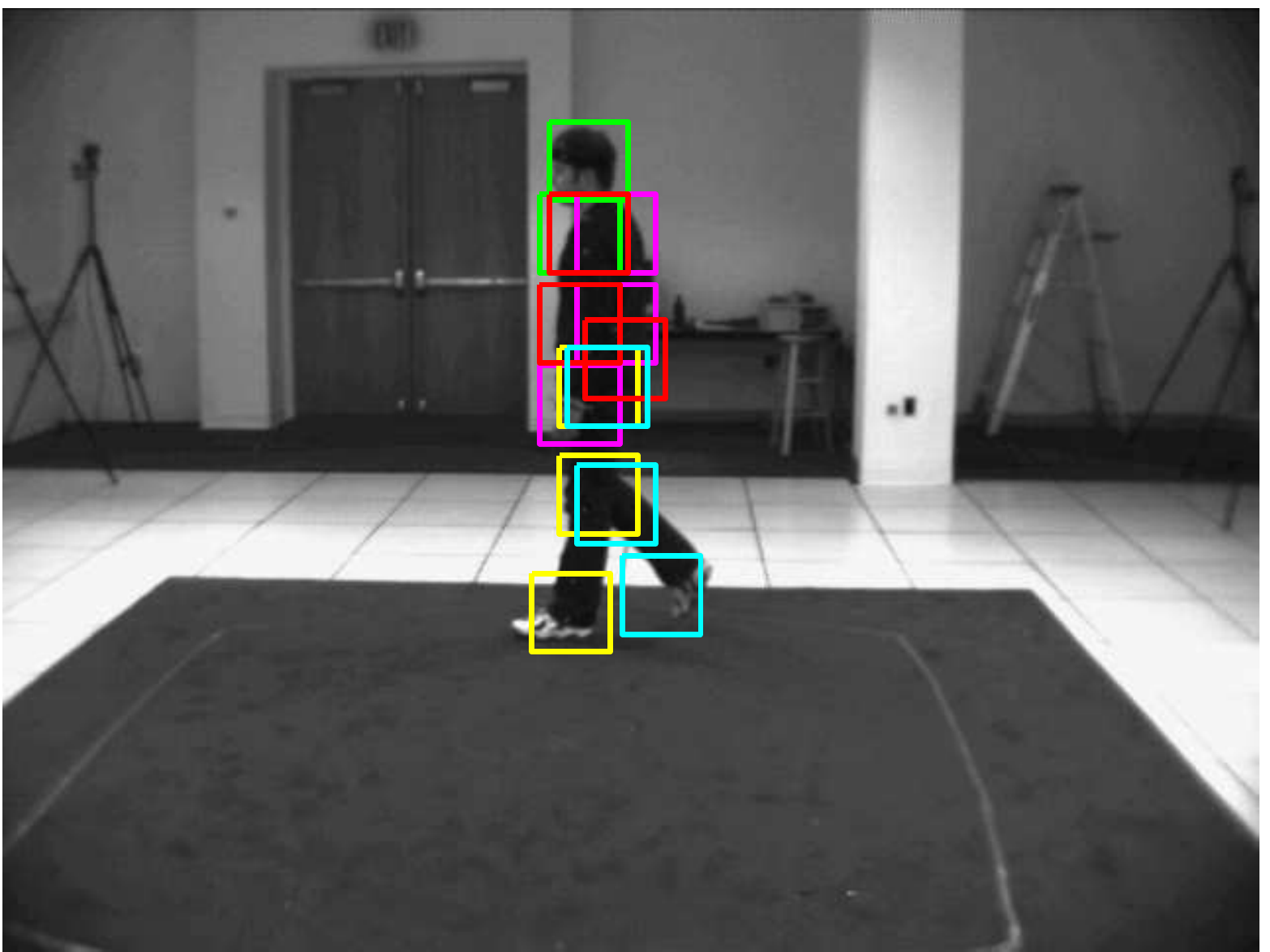} & \includegraphics[width=0.24\linewidth]{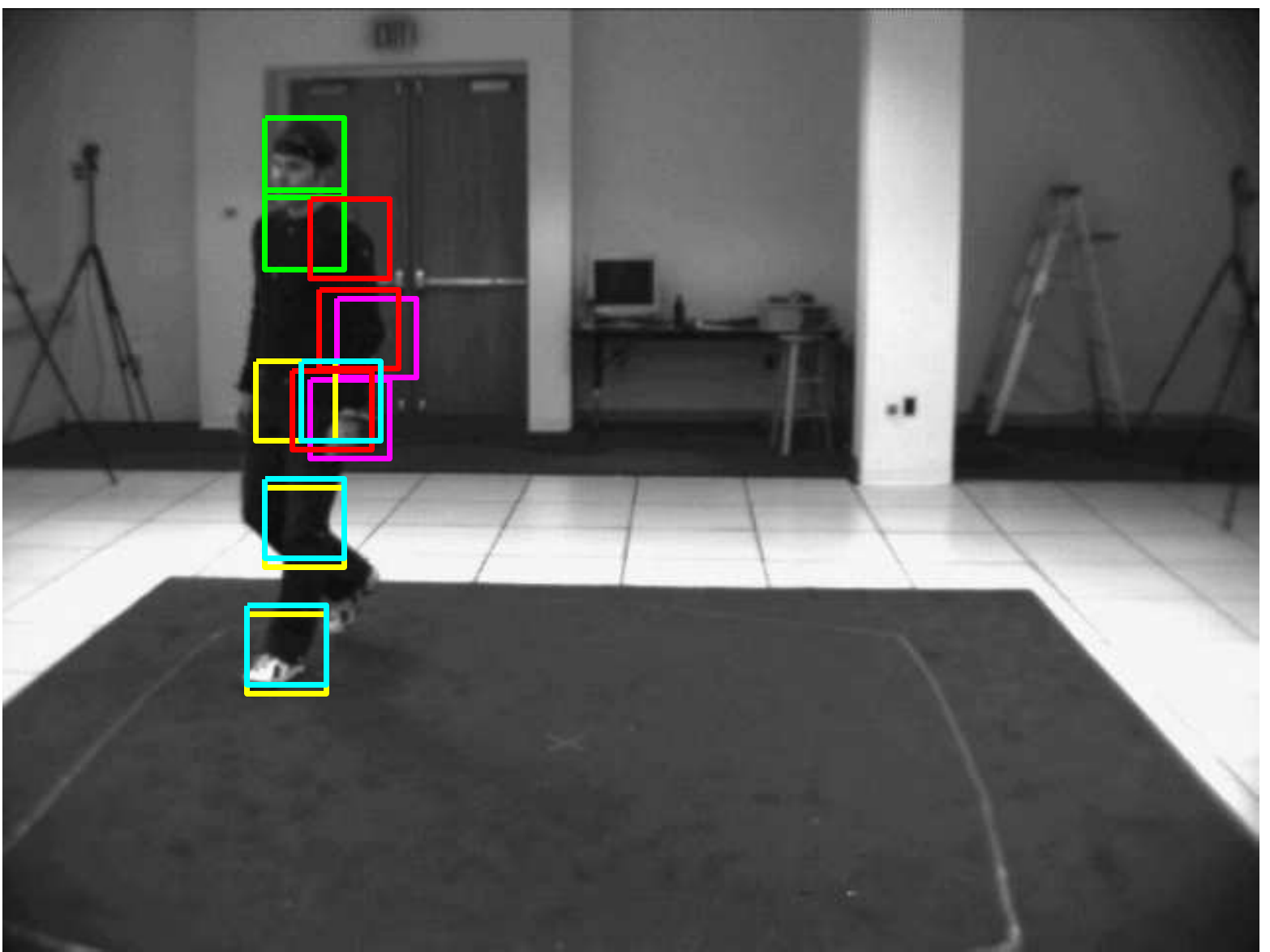} &
\includegraphics[width=0.24\linewidth]{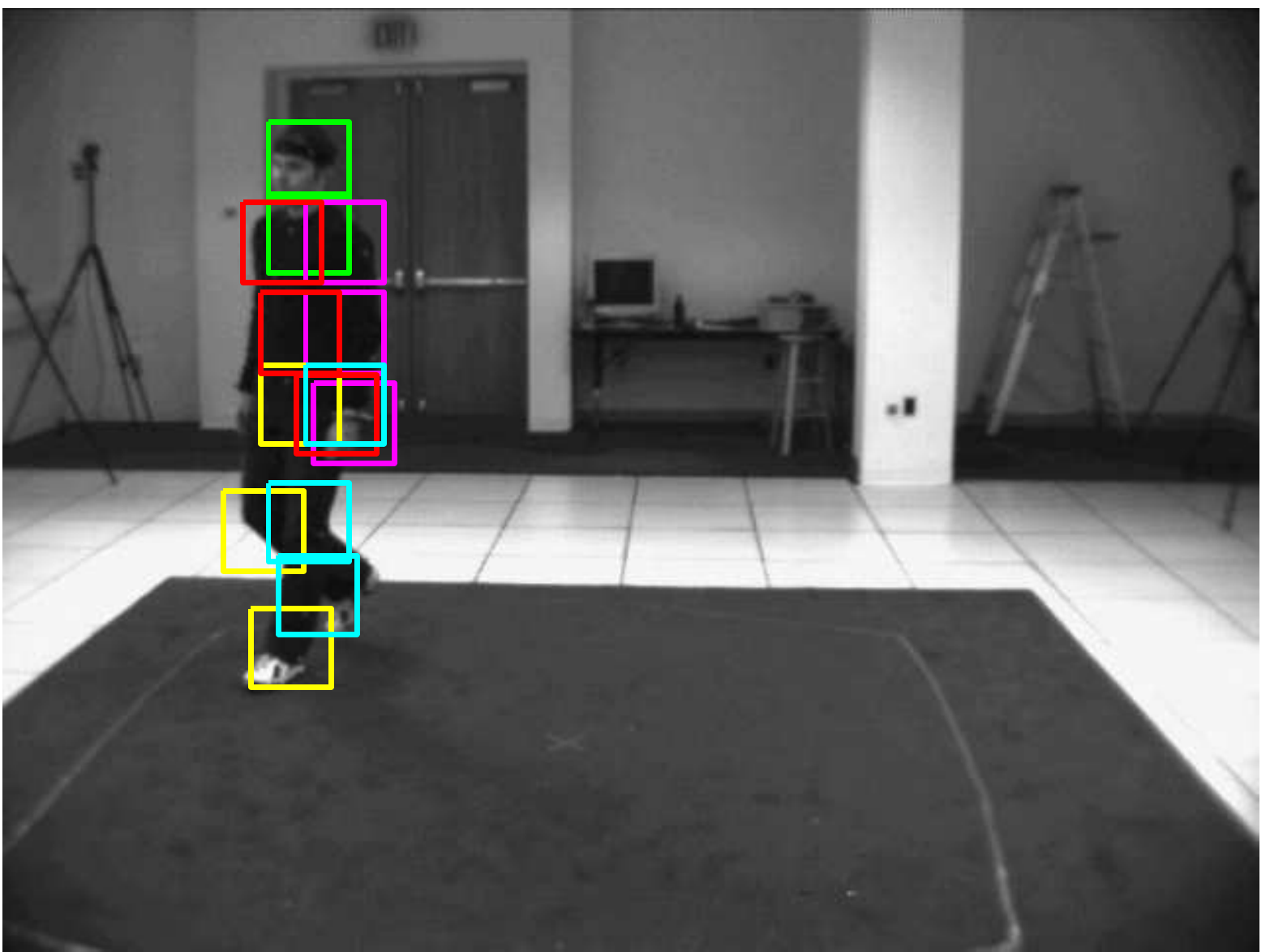} \\  \\
\end{tabularx}
\caption[Qualitative 2D pose estimation samples.]{Qualitative 2D pose estimation samples. Column ``MoP detection'' shows detected 2D body parts from the original MoP method. Column ``Our detection'' shows the enhanced 2D body part detection from our method. }
\label{fig:qualitative_results}
\end{figure*}

In our solution, we introduce another feature cue (extracted human blobs in current experiments). In this way, not only the localization could be verified by two features, but also we are able to optimize the target in a global way. This is due to the newly introduced feature cue gives global description. The following are main module that are incorporated:
\begin{itemize}
\item we keep response scores from all mixtures of the current body part for later use. Instead of selecting the mixture with the maximum response in~\cite{cvprYangR11APEMOP}, we select pre-defined set of body part candidates whose overlap with another feature are over a certain ratio and put them in a candidate list, and calculate the best configuration whose overlap of two features are the maximum. The reason that we select a \emph{pre-defined} set of body part candidates is that if we consider all body limbs, the calculation number might be too much ($5*5*5*6*6*5*6*6*5*6*6*5*6*6$ in our method, where we use fourteen body parts, with five or six mixture for each body part) and redundant because not all body limbs are possible to have double counting problem. The body parts that are possible to have double-counting problems are: two elbows (left and right), two hands, two knees and two legs. So we \emph{pre-define} a set of body parts that could be optimized.
\item we add a mixture-selection module, where any mixture that has a overlap ratio of over a certain threshold ( $thresh2 = 0.2$ in our experiment, due to the noisy extracted silhouettes) are considered as a candidate mixture that can pass messages to it parent.
\item we optimize the position among all kept candidate positions for a body part by fixing all other body part positions. If the overlap of two feature cues at a pixel location is within a interval ($thresh1 = 0.5$ in our experiment), for all pair of body part that might encounter double counting problem (that is, two elbows, including the left elbow and the right elbow, two hands, two knees and two legs), we check if they overlap. If they do, there is a possibility that they are double-counted, then they are added to the candidate list for local optimization.
\end{itemize}

Note that, in our experiments parameters are set by experience. It is also straightforward to acquire them from training data. For example, we can calculate all overlap ratios between bounding boxes of training body parts and extracted human blobs, fit Gaussian distribution, and take the mean of the fitted Gaussian as $thresh2$. Body part localization results are shown in figure~\ref{fig:qualitative_results}. From the figure, we can see that double counted body parts are correctly localized after optimization. 

\subsection{3D Pose Estimations}

\begin{figure}[!ht]
\begin{center}
\begin{tabular}{c}
\includegraphics[width=0.6\linewidth]{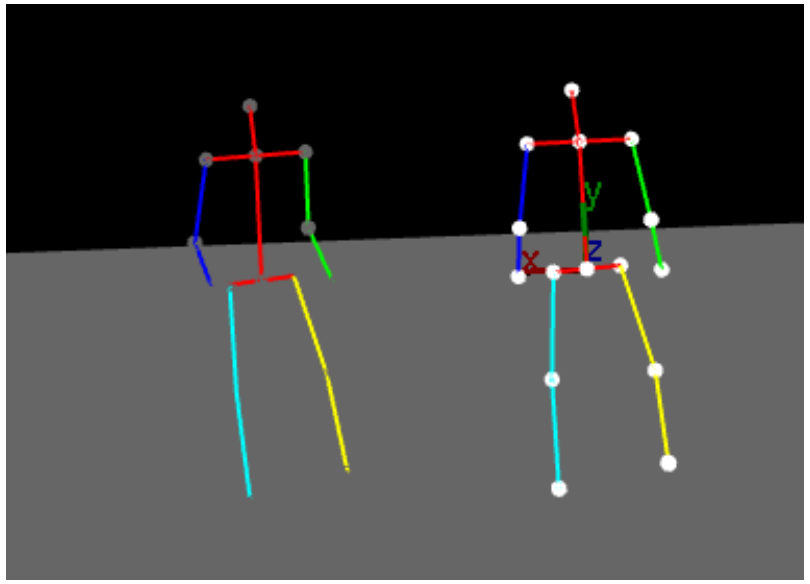}\\
\includegraphics[width=0.6\linewidth]{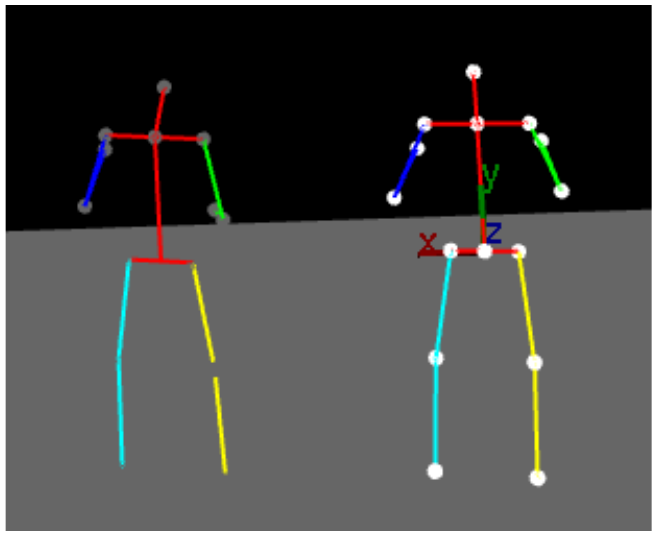}\\
\includegraphics[width=0.6\linewidth]{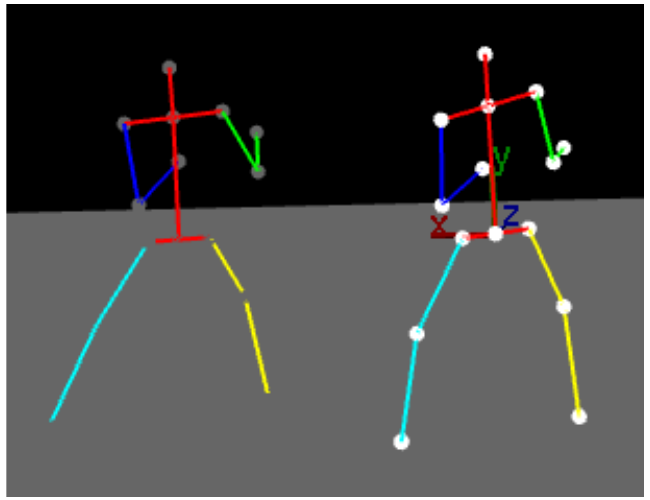}\\
\end{tabular}
\end{center}
\caption{Examples of visualized 3D pose estimation. The first row is a frame from actor S1 performing walking. The second row is a frame from actor S1 performing box. The third row is a frame from actor S2 performing box. The stick figure on the left is the ground truth data, and the stick figure on the right is the estimated 3D pose from localized 2D body part positions. }
\label{fig:visWalking}
\end{figure}
We further feed enhanced 2D body part locations to pose estimators and 3D poses are predicted. For each experiment, we train a set of Gaussian processes with Squared Exponential covariance matrix with the training set, the proposed 2D body part detectors are applied on test images, and detected 2D body parts are fed to the trained Gaussian process regressors to get 3D pose estimations. Here we show some visualizations of 3D pose estimations. Figure~\ref{fig:visWalking} shows examples from walking and box actions. 

\begin{figure}[!ht]
\begin{center}
\begin{tabular}{c}
\includegraphics[width=0.9\linewidth]{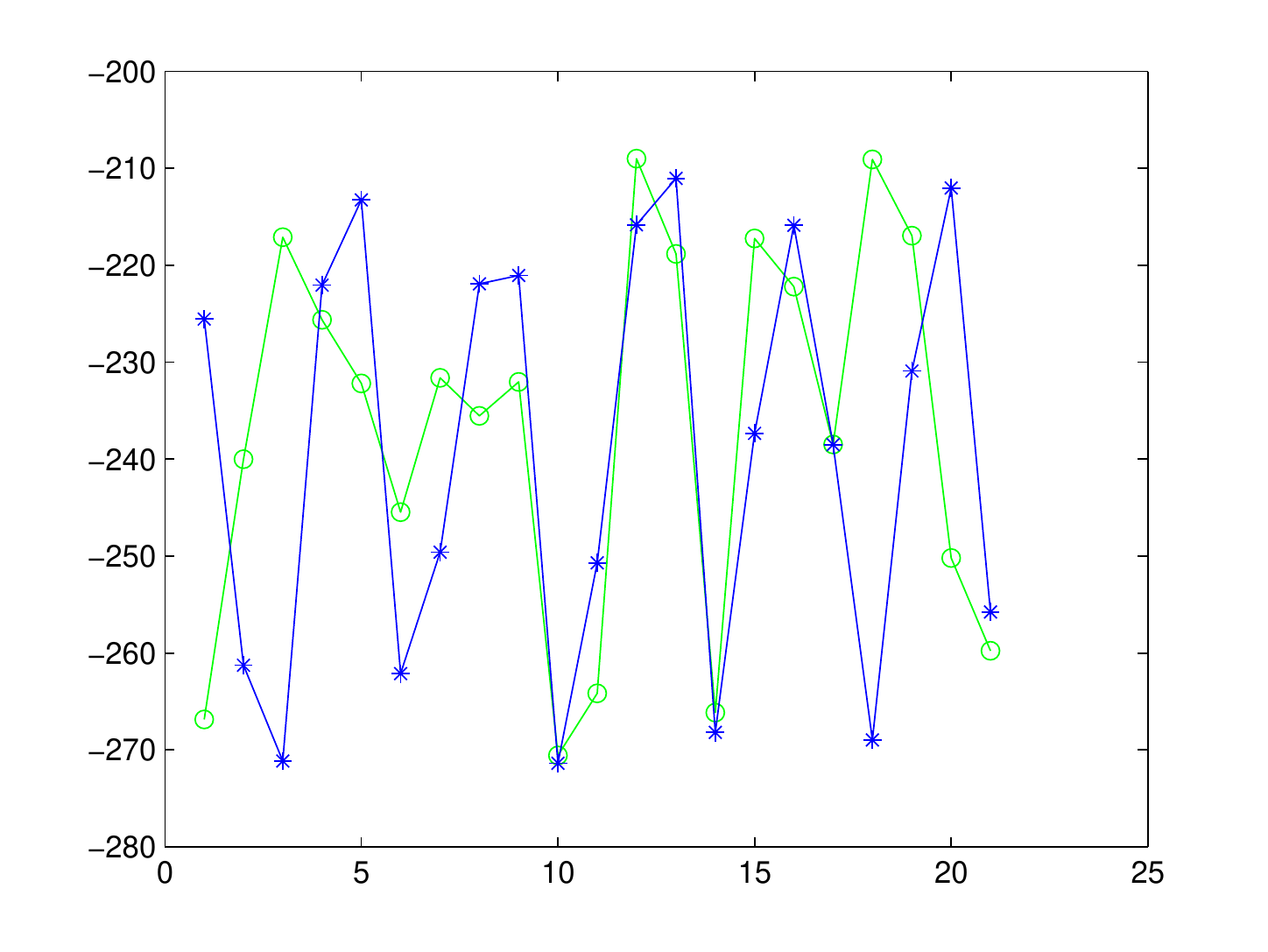}\\
\includegraphics[width=0.9\linewidth]{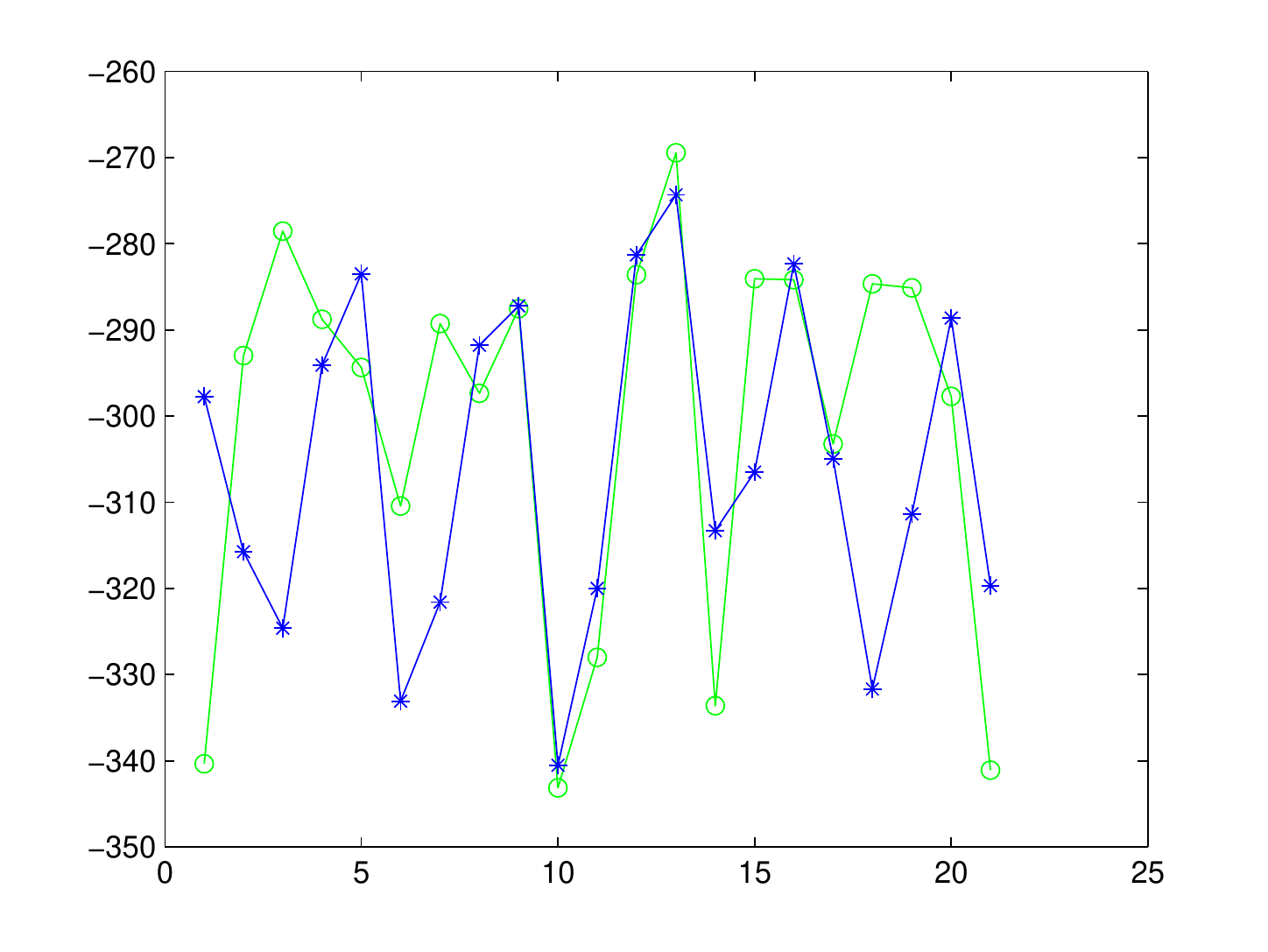}
\end{tabular}
\end{center}
\caption{The first dimension (values along axis x) of the left elbow position estimations (in green) and ground truth joint positions (in blue). The first figure is a frame from actor S2 performing walking. The second figure is a frame actor S2 performing walking. The x axis denotes frame id, ranging from $1$ to $21$. The y axis denotes values of the first dimension from 3D joint positions and its unit is milimeter.}
\label{fig:leftElbowWalkingS1}
\end{figure}
To have a qualitative comparison, we show in figure~\ref{fig:leftElbowWalkingS1} 3D joint positions of the ground truth pose and the estimated 3D pose. Both figures shows values from the first dimension of 3D joints. The figure in the first row is from the left elbow of the actor ``S2'' performing ``Walking'' and the figure in the second row is from the left hand of the actor ``S2'' performing ``Walking''.

One direct application of the proposed method is for controlling 3D poses of avatar. Motion capture systems usually requires invasive body markers. While in our pipeline, performers are able to get rid of invasive body markers once training 3D poses are attained. What's more, we only need image sequences from one single view point.

\section{Conclusions and Discussions}\label{sec:conclusions}


In this paper, we design an algorithm to enhance the performance of 2D body part localization based on Mixture of Parts models which recently achieved good performances in 2D body part localization. Later on, we take the estimated poses as an input to estimate 3D poses. We validate our method in two ways: 2D body part localization visualized results and 3D pose estimation errors. One interesting further work is to incorporate physical constraints into 3D human model so we can optimize 2D body parts accordingly. We are also interested into validate this method on other public data set, like YouTube data set where 3D pose ground truth are not provided.


\end{document}